\documentclass[journal]{IEEEtran} 
\usepackage[pdftex]{graphicx,graphics}
\usepackage{amsmath,amsfonts}
\usepackage{algorithmic}
\usepackage{algorithm}
\usepackage{array}
\usepackage[subrefformat=parens]{subcaption}
\usepackage{textcomp}
\usepackage{stfloats}
\usepackage{url}
\usepackage{verbatim}
\usepackage{bm}
\usepackage{cite}

\begin{document}

\title{Power in Numbers: Primitive Algorithm for Swarm Robot Navigation in Unknown Environments}

\author{Yusuke Tsunoda, Shoken Otsuka, Kazuki Ito, Runze Xiao, \\ Keisuke Naniwa, Yuichiro Sueoka, Koichi Osuka
\thanks{}
\thanks{}}

\markboth{}%
{Shell \MakeLowercase{\textit{et al.}}: A Sample Article Using IEEEtran.cls for IEEE Journals}

\IEEEpubid{}

\maketitle

\begin{abstract}
Recently, the navigation of mobile robots in unknown environments has become a particularly significant research topic.  
Previous studies have primarily employed real-time environmental mapping using cameras and LiDAR, along with self-localization and path generation based on those maps.
Additionally, there is research on Sim-to-Real transfer, where robots acquire behaviors through pre-trained reinforcement learning and apply these learned actions in real-world navigation. 
However, strictly the observe action and modelling of unknown environments that change unpredictably over time with accuracy and precision is an extremely complex endeavor.  
This study proposes a simple navigation algorithm for traversing unknown environments by utilizes the number of swarm robots.
The proposed algorithm assumes that the robot has only the simple function of sensing the direction of the goal and the relative positions of the surrounding robots.
The robots can navigate an unknown environment by simply continuing towards the goal while bypassing surrounding robots.
The method does not need to sense the environment, determine whether they or other robots are stuck, or do the complicated inter-robot communication.
We mathematically validate the proposed navigation algorithm, present numerical simulations based on the potential field method, and conduct experimental demonstrations using developed robots based on the sound fields for navigation.

\end{abstract}

\begin{IEEEkeywords}
Swarm robotics, navigation, unknown environment, potential field, self-organization
\end{IEEEkeywords}

\section{Introduction}
The navigation of robots in unknown environments has garnered significant global interest.
An unknown environment refers to a dynamic and unpredictable setting, such as a disaster site or the lunar surface, where the environment through means governed by physical laws but are difficult to anticipate \cite{open}.
In such environments, when robots are employed as substitutes for humans in search and rescue operations or transport tasks, a sophisticated navigation capability is essential to enable the unit to reach the operational area from the initial position.
This capability would generally equip the robot with the ability to sense difficult-to-traverse areas, generate a path to avoid them, and move along that path.
Various approaches have been explored in prior research on robot navigation in unknown environments to achieve this functionality.
For example, in simultaneous localization and mapping (SLAM), robots are equipped with external sensors, such as cameras or LiDAR, allowing them to observe and model the surrounding environment. 
Consequently, robots generate a path to the goal, avoiding obstacles along the way \cite{thrun2005multi,ng2007performance,malavazi2018lidar,1638022}.
Another approach, inspired by insect behavior, is the bug algorithm, where the robot alternates between moving straight towards the goal and following the wall of an obstacle to generate a path \cite{lumelsky1987path,4209200}.
In this method, the robot generates a path by alternating between moving straight towards the goal and following the wall of obstacles.
Research efforts that have investigated the application of multi-robot systems for navigation in unknown environments, including studies that apply the SLAM method to swarm robot systems \cite{atanasov2015decentralized,10.3389/frobt.2021.618268,kegeleirs2021swarm} and research on spatial exploration by small drone swarms based on the bug algorithm \cite{mcguire2019minimal}, have demonstrated that such systems exhibit high efficiency and superior fault tolerance.
Another prominent area of research is Sim-to-Real transfer, where robots acquire behaviors in a physical simulation environment using reinforcement learning, and the learned model is applied to actual mobile robots for real-world testing \cite{lee2020learning,doi:10.1126/scirobotics.abk2822,zheng2021learning}.

However, these methods are based on the assumption that the environmental data obtained from external sensors is reliable, but this information is not always accurate in unknown environments.
This is because external sensors, such as cameras and LiDAR, can capture the shape of the terrain but cannot detect the physical properties (texture information) of the terrain.
For example, what appears to be dry ground may, in fact, be soft or muddy terrain, or there may be hidden depressions beneath fallen leaves.
In an unknown environment, the true state of the environment cannot be confirmed unless the robot physically interacts with the terrain by traversing it.
It is difficult to predict environmental changes accurately in an unknown environment, as it would contradict the principles of causality.
Moreover, it is difficult to accurately model the dynamic interaction between the robot and the environment in a physical simulation environment.
Therefore, conventional methods that aim to equip a single robot with sophisticated sensory and computational capabilities are insufficient for solving the navigation problem in unknown environments, necessitating a dramatic shift in the approach followed to address the issue.

\begin{figure}[t]
    \centering 
     \includegraphics[width=0.9\columnwidth]{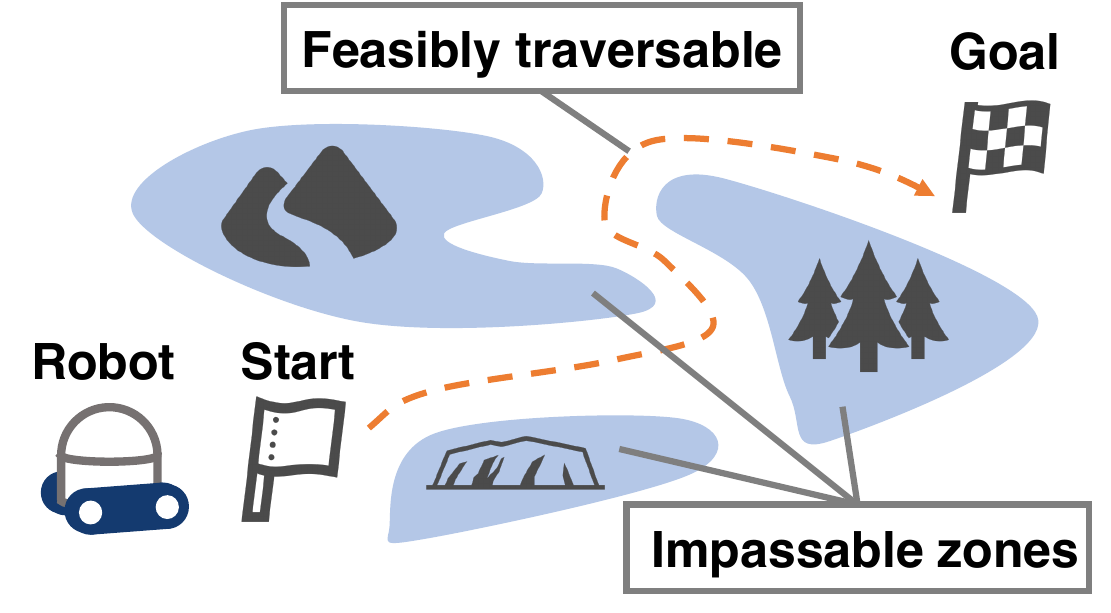}
     \caption{Diagram of problem definition for unknown environment navigation. The robot moves from the start point to the goal on a 2D plane where several areas through which the robot cannot move (impassable zones) randomly. However, we assume the existence of at least one path from the start point to the goal. Note that the robot has no prior awareness of the impassable zones.}
     \label{fig:problem_description}
\end{figure}

\begin{figure}[t]
    \centering 
     \includegraphics[width=0.9\columnwidth]{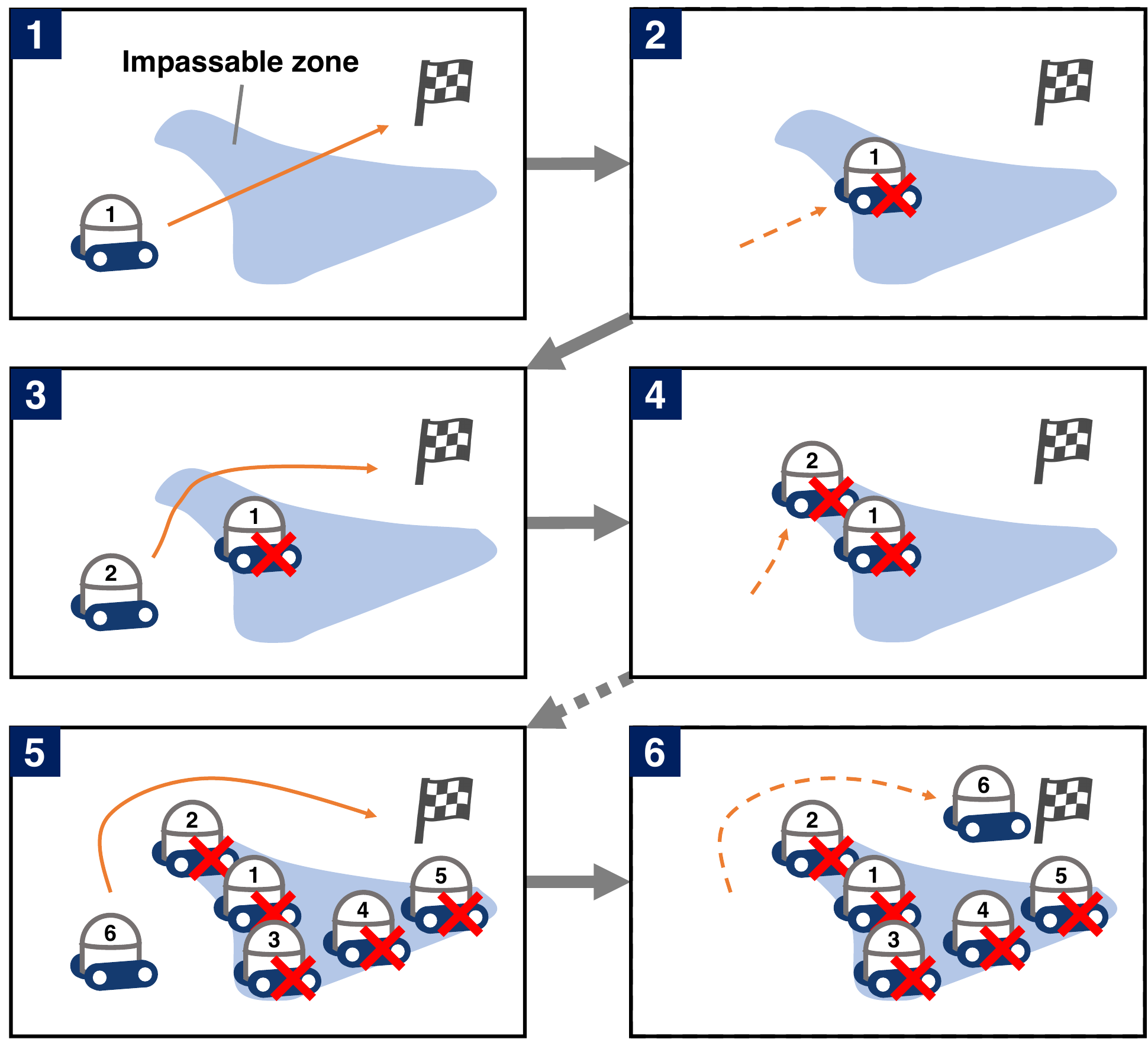}
     \caption{Schematic diagram of the proposed algorithm, BYCOMS (BYpassing COmpanions Method for Swarm robots navigation). The robots move toward the goal one at a time. The first robot gets stuck at the boundary of the impassable zone, and the following robots bypass the stuck robot and head toward the goal.}
     \label{fig:BYCOMS_overview}
\end{figure}

Therefore, we consider the eventuality that the robots will get stuck in unknown environments as inevitable, and aim to solve this problem with a primitive algorithm-based swarm robot system that utilizes several robots.
This study presents a primitive algorithm for a swarm robot system that significantly simplifies the navigation of unknown environments by leveraging the advantages of deploying multiple robots.
As illustrated in Fig.~\ref{fig:problem_description}, we address a navigation problem in a 2D plane where at least one robot from the swarm must reach a specified goal, despite the presence of randomly distributed impassable zones.
We assume the existence of at least one path to the goal that can feasibly be traversed by the robot.
Note that the robots have no prior awareness the positions and shapes of the impassable zones (depicted in the blue in Fig.~\ref{fig:problem_description}).
These zones include obstacles, such as walls, terrain undulations, mud, pits, puddles, or steps.
Each robot is assumed to only know the direction of its destination and is equipped with simple local sensing and control capabilities, specifically to enable the unit to move towards the goal while bypassing surrounding robots.
As shown in Fig.~\ref{fig:BYCOMS_overview}, robots initially deployed into the unknown environment may become stuck at the boundaries of the impassable zones.
Robots deployed subsequently bypass these stuck robots while continuing towards the goal.
By following the boundary paths created by the stuck robots, some robots eventually navigate through the unknown environment and reach the goal.
We refer to this proposed method as BYCOMS (BYpassing COmpanions Method for Swarm robots navigation).
The key feature of this approach is that it considers the fact that robots get stuck in an unknown environment as an inevitable eventuality, and it is only viable in a multi-robot swarm system. 
Moreover, each robot programmed based on the proposed method has the significant advantage of not requiring any environmental sensing, detection capabilities to evaluate whether it is stuck or a peer robot is stuck, or even inter-robot complicated communication. 
Conventional studies of swarm robot systems have shown that task efficiency improves with an increase in the number of robots. 
However, this also introduces challenges such as higher communication loads, synchronization issues, and increased coordination complexity\cite{mcguire2019minimal,10.1007/978-3-540-30217-9_86,1362262944632390912}. 
In contrast, the proposed algorithm avoids these complications by not requiring complex adjustments when scaling up the number of robots. Instead, it directly enhances the task accomplishment rate, enabling robots to navigate more complex and unknown environments effectively.
In related work of this study, prior research on the bug algorithm has mathematically proven the feasibility of a robot reaching its goal by following the edge of an obstacle \cite{lumelsky1987path,4209200,doi:10.1177/0037549720930082}; however, this approach assumes that the robot can sense the side of the obstacle and move along it to detect the perimeter.
This differs from the capabilities defined for the robots in this study. 
Furthermore, in this study, areas that robots should avoid, such as walls or obstacles, are more broadly defined as impassable zones where the robot becomes stuck and immobile, thereby addressing a more generalized navigation problem.
Additionally, there is research that utilizes robots that are sacrificed by falling into pits or off ledges to achieve the cooperative traversal of a swarm with simple control \cite{sugawara2018casualty}.
However, no mathematical proof has been provided regarding the feasibility of the navigation system ability in enabling the robot to reach its goal.

First, we formulate the swarm robot navigation problem and mathematically demonstrate the goal reachability of swarm robots using the BYCOMS method.
Subsequently, we simulate the proposed method by representing the robots' avoidance behavior as movement along a virtual potential field created by surrounding robots\cite{fujisawa2008communication,sugawara2004foraging}.
Thereafter, we examine two aspects through simulation: changes in the avoidance radius of robots and the probability of reaching the goal and robustness to noise in local sensing.
Finally, we employ sound waves as the physical quantity to create the potential field and develop robots that navigate according to the sound field using speakers and microphones \cite{YuichiroSUEOKA202120-00280,tsunoda2019experimental,Yusuke_Tsunoda2023}.
Through the experiments, we demonstrate the practical feasibility of our proposed method.
The structure of this paper is organized as follows. 
Section \ref{sec:Problem} discusses the problem setting of swarm robot navigation in unknown environments and introduces the proposed algorithm. 
Section \ref{sec:Theorem} mathematically proves the goal reachability of robots based on the proposed method.
Section \ref{sec:Sim} describes the implementation of the proposed algorithm based on the modification of a virtual potential field. 
In Section \ref{sec:sim_result}, we evaluate the performance of swarm robot navigation through numerical simulations. 
Section \ref{sec:Expe} presents the robot implementation using sound fields and examines the feasibility of swarm robot navigation. 
Finally, Section \ref{sec:Conc} provides a summary of the study and discusses the scope for future research.


\section{Proposal of BYCOMS}
\label{sec:Problem}
First, we outline the problem setting of swarm robot navigation in unknown environments addressed in this study and provide an overview of the proposed navigation method.

\subsection{Problem definition and control objective}
\label{sec:Purpose}

\begin{figure}[t]
    \centering 
     \includegraphics[width=\columnwidth]{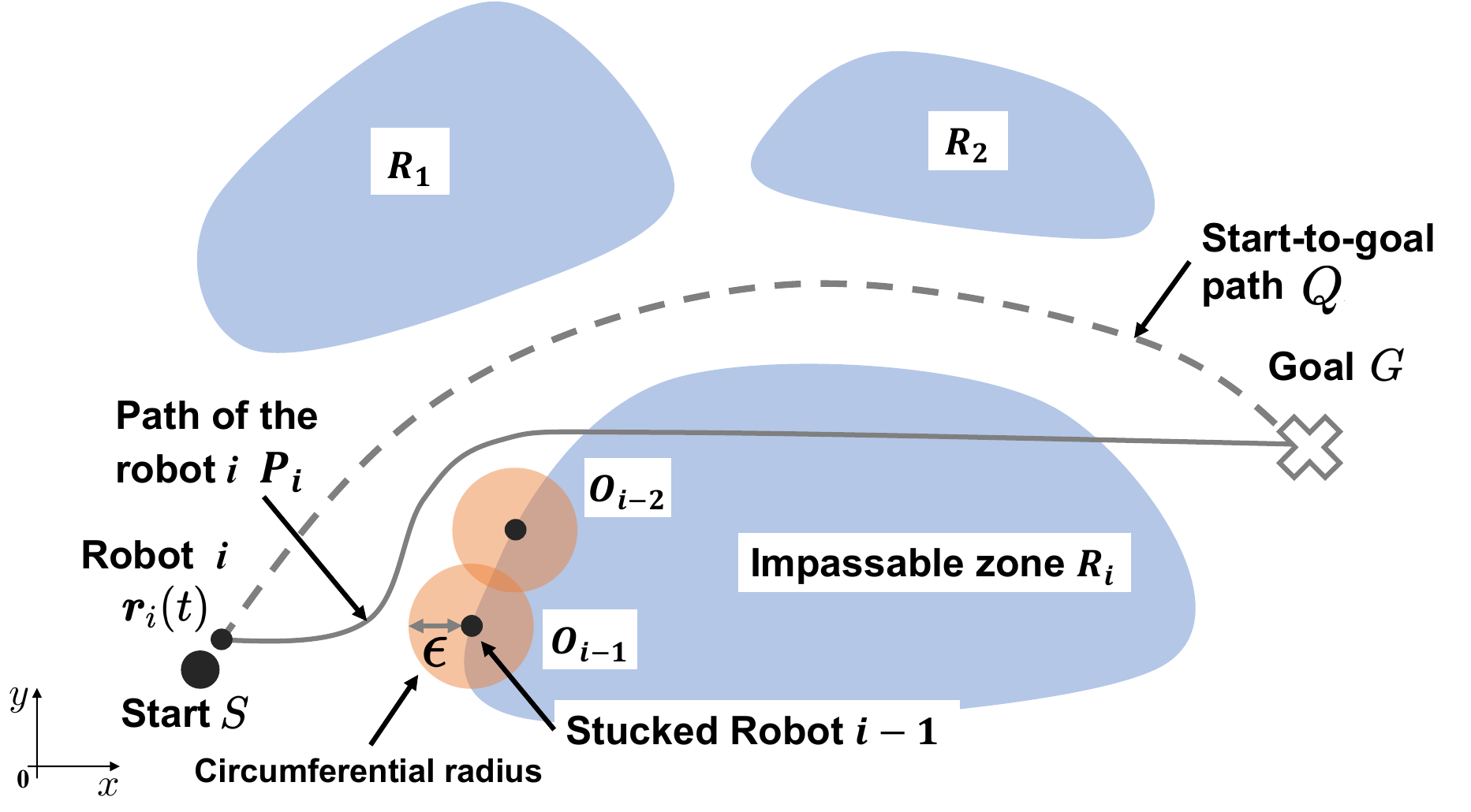}
     \caption{
        Schematic diagram of the mathematical formulation of the guidance problem. In a 2D plane, the start point is denoted as $S$, the goal as $G$, and the impassable zone for robot $i$ as a closed region $R_i~(i=1,\cdots,n)$. The position of robot $i~(i=1,\cdots,N)$ is represented by $r_i$, and when robot $i$ gets stuck, it generates a circular closed region with radius $\epsilon$, denoted as $O_i$. The path that a robot can take from the start to the goal is represented by $Q$, while the path $P_i$ is the route for robot $i$ that avoids passing through the regions $O_{i-1}, O_{i-2}, \cdots$, generated by the previous robots.
        }
     \label{fig:math_model}
\end{figure}
\begin{figure}[t]
    \centering 
     \begin{minipage}[t]{0.85\columnwidth}
      \includegraphics[width=\columnwidth]{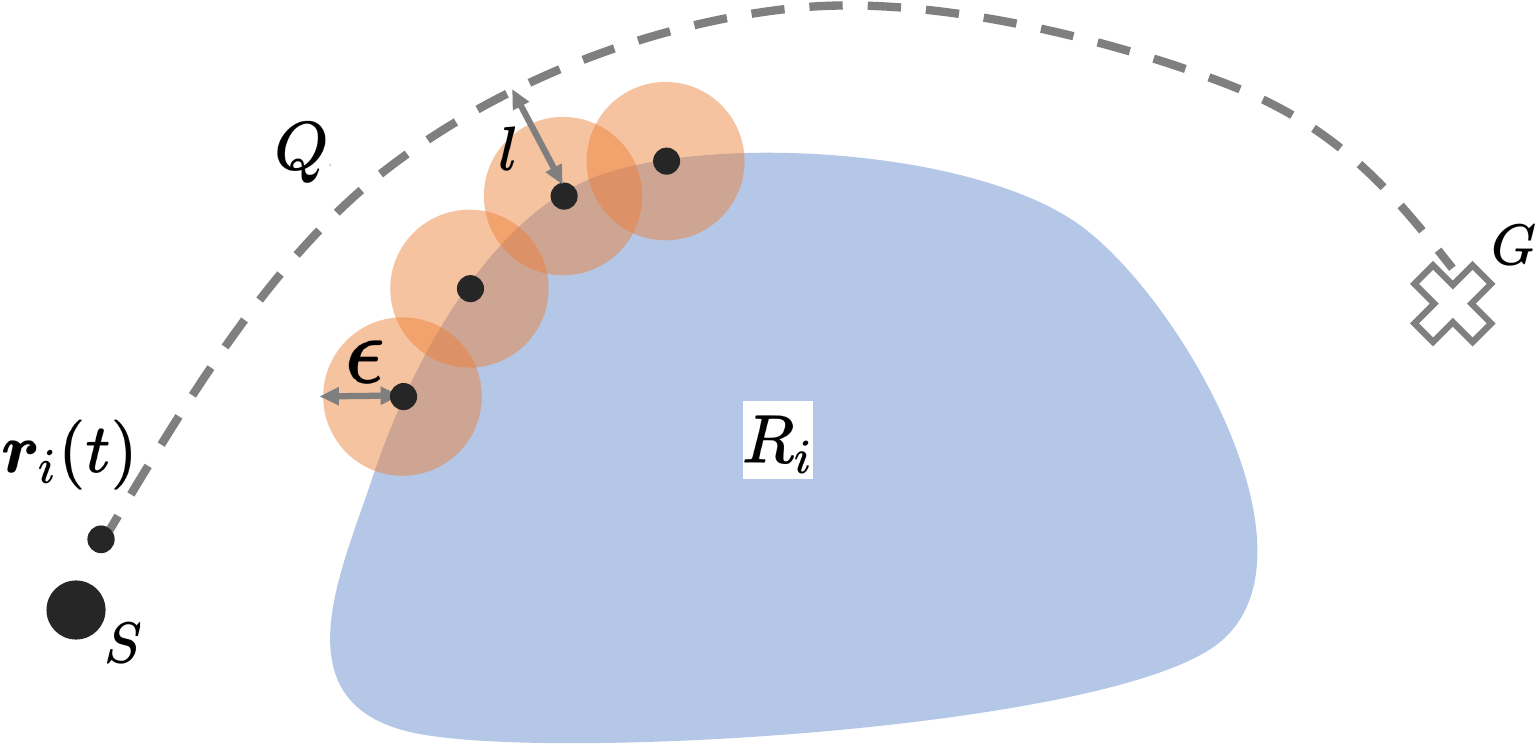}
      \subcaption{
        Relationship between the passable path $Q$ from the start point to the goal, the minimum distance $l$ from the stuck robot, and the robot's circumferential radius $\epsilon$. By setting $\epsilon$ such that $\epsilon < l$, the path $Q$ is maintained.
        }
      \label{fig:method_l_eps}
     \end{minipage}
     \begin{minipage}[t]{0.8\columnwidth}
      \includegraphics[width=\columnwidth]{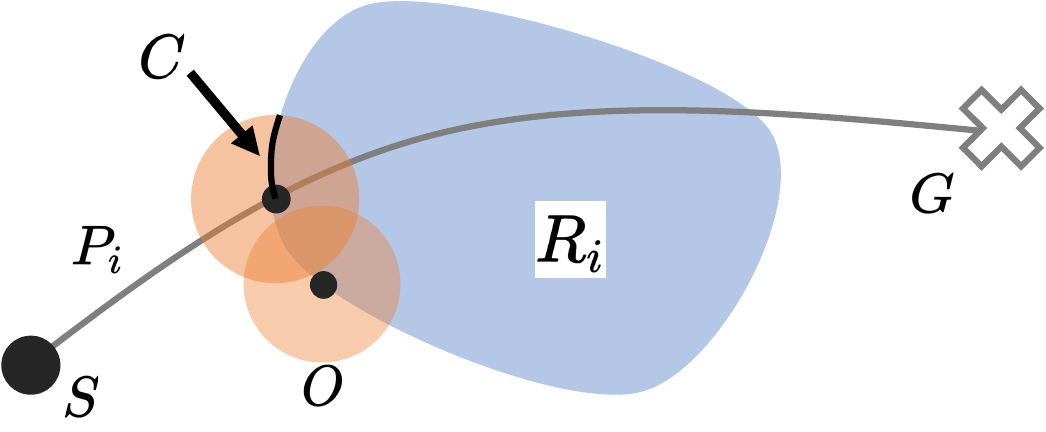}
      \subcaption{
        When robot $i$ gets stuck, a non-zero length boundary line $C$ is newly included in $O$.
        }
      \label{fig:method_c}
     \end{minipage}
     \begin{minipage}[t]{0.8\columnwidth}
        \includegraphics[width=\columnwidth]{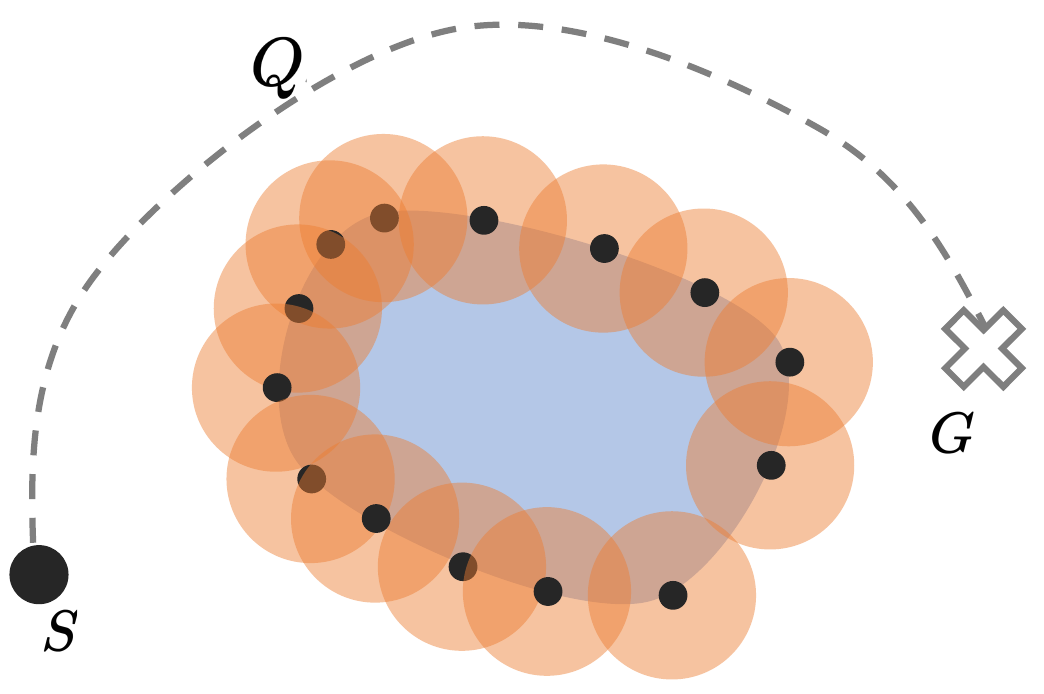}
        \subcaption{
            If the robots continue to get stuck, the boundaries of $R_1, R_2, \cdots , R_n$ will all be encompassed within $O_1, O_2, \cdots$ by a finite number of robots.
            }
        \label{fig:method_com}
       \end{minipage}
    \caption{Overview charts in BYCOMS reachability considerations}
    \label{fig:method_outline}
\end{figure}

Figure~\ref{fig:math_model} illustrates the outline of the navigation problem considered in this study. 
As shown in Fig.~\ref{fig:math_model}, we consider the navigation of $n \in \mathbb{N}$ robots in a two-dimensional plane $W \in \mathbb{R}^\mathrm{2}$ within an unknown environment. 
The plane $W$ contains a finite number of static closed zones $R_1, R_2, \cdots ,R_n (n \in \mathbb{N})$ with finite perimeters, within which the robots cannot move (hereafter referred to as impassable zones).
In $W$, there exist two points outside the impassable zones: a start point $S = [x_s, y_s]^\mathrm{T}$ and a goal $G = [x_g, y_g]^\mathrm{T}$. 
At the beginning of navigation, it is assumed that at least one path $Q$ from the start to the goal does not pass through the closed zones $R_1, R_2, \cdots ,R_n$.

Here, the control objective is for at least one of the $N$ robots starting from $S$ to reach the goal $G$ within a finite time. 
We set the position of a robot $i~(i=1,2,\cdots,N)$ at time $t$ as $r_i(t) = [x_i(t), y_i(t)]^\mathrm{T}$, and then the control objective is formulated as follows:
\begin{eqnarray}
\exists i  ~ : \lim_{t \to \infty} \|\bm{r}_{i}(t) - \bm{G}\| = 0.
\end{eqnarray}

\subsection{Robot algorithm}
\label{sec:Algorithm}

Subsequently, we describe the proposed algorithm. 
Each robot is sequentially placed one at a time at the start location with a sufficiently large time interval, and then begins the navigation. 
The robots have the following two functions:

1. Generate a path from the current location to the goal in real-time, avoiding a circular zone with radius $\epsilon$ (circumferential radius) centered at the positions of all other robots.

2. Move along the path generated in the above function.
Here, the circular zones with the radius $\epsilon$ generated by each robot are represented as virtual potential fields.
Specifically, each robot generates a path to the goal that avoids the virtual potential fields created by surrounding robots and moves along this path.
The specific implementation of the algorithm based on virtual potential fields is explained in Section~\ref{sec:Potential}.

Thereafter, we explain the overall algorithm for the system. 
When the robot's position is within $R_1,R_2,\cdots ,R_n$, the robot gets stuck and creates a circular closed region of radius $\epsilon$.
As shown in Fig.~\ref{fig:math_model}, the $i$-th robot, $r_i$, obtains the coordinates of other robots and the goal, and generates a path $P_i$ to the goal that avoids circular closed zones $O_1, O_2, \cdots ,O_{i-1}$ centered on the coordinates of other robots, with the circumferential radius $\epsilon$.
Because the robots cannot observe the terrain, they initially gets stuck on the path to the goal. 
As this process continues with multiple robots, the impassable zones are filled by stuck robots, allowing subsequent robots to bypass the earlier robots. 
Consequently, the robots can eventually avoid the impassable zones and reach the goal.

\section{Proof of goal reachability}
\label{sec:Theorem}
We mathematically prove the reachability of the robot's goal when the proposed method is applied.
Based on the above assumptions, we prove the following propositions:

{\bf [Proposition]}

At least one of the $N$ robots in a swarm based on BYCOMS can reach the goal within a finite time in any unknown environment.

{\bf [Proof]}

From the assumptions above, we can consider that only one robot is moving at any particular time. 
The first robot $i=1$ generates and moves along a path $P_1$ as there are no other robots in the environment.
For the robot $i$ that can identify a path $P_i$, the robot could potentially generate one of two kinds of paths: when $P_i$ does not intersect with $R_1, R_2, \cdots, R_n$ and when it does.
First, if $P_i$ does not intersect with $R_1, R_2, \cdots, R_n$, robot $i$ can reach the goal. 
If $P_i$ intersects with $R_1, R_2, \cdots, R_n$, the robot $i$ gets stuck and stops on the boundary of $R_1, R_2, \cdots, R_n$ at the point where the path $P_i$ from the start point is shortest. 
Because $r_i$ is on the boundary of $R_1, R_2, \cdots, R_n$, setting $\epsilon$ such that $\epsilon < l$, where $l$ is the minimum distance between $R_1, R_2, \cdots, R_n$ and $Q$, ensures that $Q$ does not pass through $O_1, O_2, \cdots, O_i$ (Fig.~\ref{fig:method_l_eps}). 
Therefore, robot $i$ can always generate a path $P_i$ under this condition because $Q$ exists as a path.

When robot $i$ gets stuck, a non-zero length boundary $C$ of $R_1, R_2, \cdots, R_n$ is added to $O_1, O_2, \cdots, O_i$ (Fig.~\ref{fig:method_c})
If the robots sequentially deployed from the start point continue to get stuck, the total boundary length of $R_1, R_2, \cdots, R_n$ is finite; thus a finite number of robots cover the entire boundary of $R_1, R_2, \cdots, R_n$ within $O_1, O_2, \cdots$. 
We refer to this state as the comprehensive state (Fig.~\ref{fig:method_com}).
When the system is in the comprehensive state, if a path to the goal exists that does not pass through $O_1, O_2, \cdots$, this path does not intersect with $R_1, R_2, \cdots, R_n$, allowing the robot to reach the goal without getting stuck. 
Because at least $Q$ exists as a path in this state, the robot can reach the goal.
From the proof above, we can conclude that under the given conditions, the proposed navigation method ensures that one robot can reach the goal in any terrain.

\section{Simulation of BYCOMS}
\label{sec:Sim}
In this section, we implement the proposed algorithm based on the modification of potential fields and verify the feasibility of navigating unknown environments through simulations. 
Additionally, we investigate the impact of the robot's circumferential radius parameter on traversal performances, as well as the robustness against sensor noises.

\subsection{Simulation environment}
\begin{figure}[t]
    \centering 
     \includegraphics[width=0.9\columnwidth]{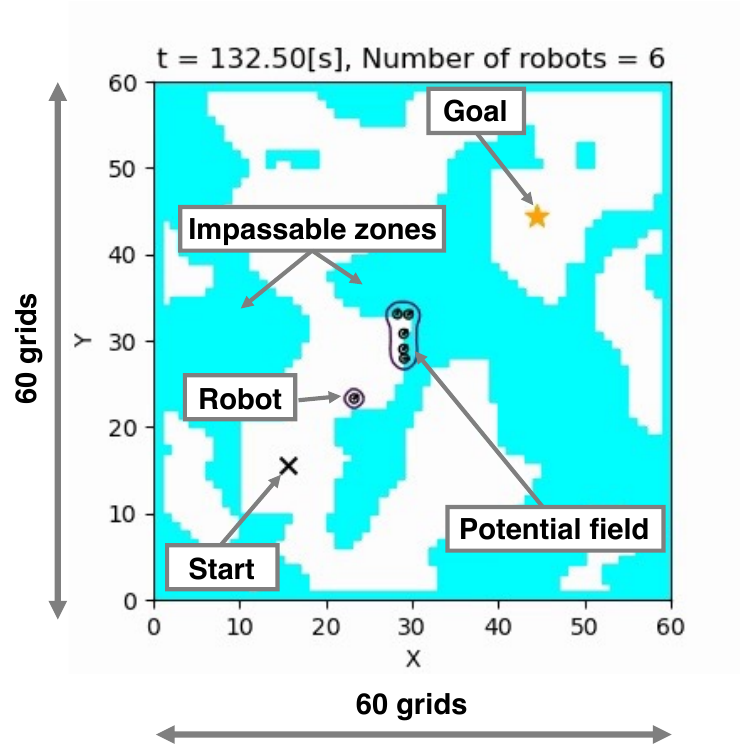}
     \caption{
     Simulation enviroment for unknown environment navigation. The simulation is conducted on a 60x60 grid in a two-dimensional plane. The start point is marked by an cross, the goal by a star, and the robots are represented as circles. The purple lines indicate potential contours, and the blue areas represent the robots' impassable zones.
     }
     \label{fig:exp_sample}
\end{figure}

An example of the navigation environment is shown in Fig.~\ref{fig:exp_sample}. 
The simulations are conducted using Python 3. 
The entire environment is a square on a 2D plane with sides of length 60 units. 
The environment is divided into a $60 \times 60$ grid, with each cell set to either a passable or impassable state for the robots. 
In Fig.~\ref{fig:exp_sample}, the light blue cells represent areas where robots cannot move, circles represent robots, the cross mark is the start point, and the star represents the goal. 
Robots are treated as points, and robot-to-robot contact is not considered. 
Each robot moves while generating a potential field.
The purple lines represent the contours of the potential fields created by the robots.
If the robot enters a light blue cell, it is considered stuck and immobilized.

The start point and goal are positioned diagonally at coordinates $[15.5, 15.5]^\mathrm{T}$ and $[44.5, 44.5]^\mathrm{T}$, respectively. 
After a robot gets stuck, a new robot is placed at the starting position, facing the goal.
The simulation ends when the robot's center coordinates are within the robot's radius from the goal, indicating the robot has reached the goal, or after $10000\textrm{s}$. 
The simulation calculation period is set to $0.1\textrm{s}$, and the value of the grid width is $1$.

\subsection{Generation of unknown environments}
To generate the terrain, we use Perlin noise, which is commonly used for creating natural and random terrain representations \cite{rose2016algorithms}. 
Perlin noise generates a two-dimensional scalar function $f(x,y)$ defined on the $xy$ plane.
Using a uniformly distributed random real number $a$ in the range of $-1$ to $1$, each grid is classified as either a traversable or non-traversable area based on the value of $f$ at the center of the grid. 
Specifically, if the value of $f$ at the center of a grid is greater than $a$, the grid is designated as a non-traversable area. 
Conversely, if the value of $f$ is less than or equal to $a$, the grid is designated as a traversable area. 
However, a $10 \times 10$ square zone centered around the start point and goal positions is always designated as a traversable area to ensure the presence of a viable path. 
To verify the existence of a path from the start point to the goal that only passes through traversable areas, we use the A* algorithm \cite{warren1993fast}. 
Only terrains where such a path exists are utilized for the simulation.

\subsection{Implementation of BYCOMS based on virtual potential fields}
\label{sec:Potential}
We implement the proposed algorithm inspired by the bug algorithm \cite{lumelsky1987path}. 
The robots employ global sensing to determine the direction to the goal and local sensing to evaluate the strength and gradient of the potential field at their current location.
Each robot generates a physical quantity with wave-like properties, such as light, sound, or radio waves, with its center as the source. 
This allows the robot to form a potential field and sense the potential fields created by other robots, guiding its movement.
The strength of the potential field decays according to the $r^2$ law with respect to the distance $r$ from the potential source. 
Specifically, we define the potential field strength for the robot positioned at $[x_r, y_r]^\mathrm{T}$ as follows:
\begin{equation}
    g(x,y) = \frac{c}{(x-x_r)^2+(y-y_r)^2},
\end{equation}
where $c$ is a constant parameter. 
In the numerical simulations described in Section \ref{sec:Sim}, the constant parameter $c$ is set to a value of $1$.
Moreover, each robot has a certain threshold $g_{th}$ for the intensity of the potential.
It evaluates the distance from surrounding robots by comparing the potential intensity $g$ it senses on the spot with this threshold value.
Specifically, the threshold $g_{th}$ adjusts the circumference radius $\epsilon$ that each robot uses to bypass other robots.

\begin{figure}[t]
    \centering 
        \includegraphics[width=\columnwidth]{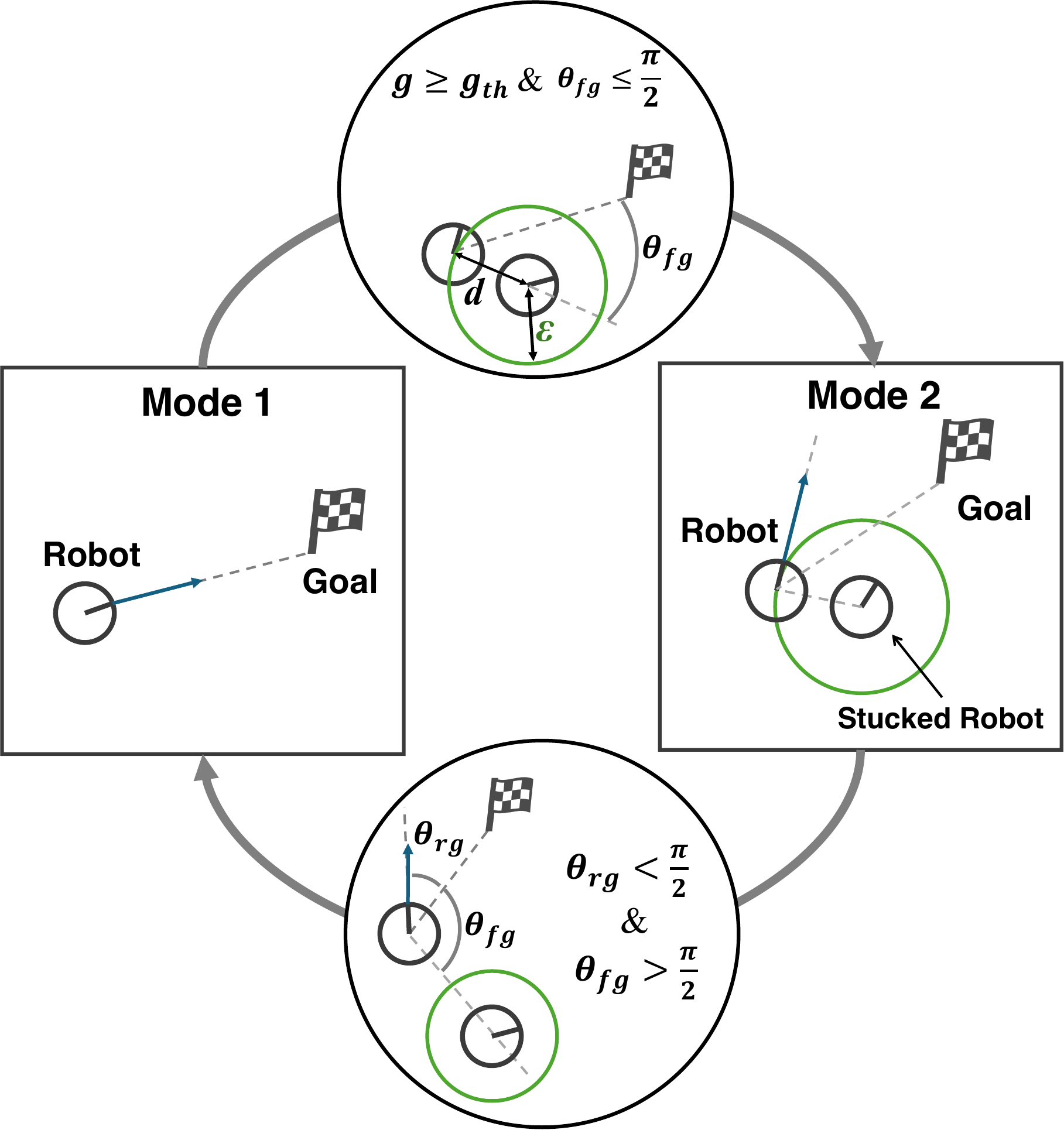}
    \caption{
    Robot algorithm based on the potential fields. The robots transition between two behavior modes: Mode 1 and Mode 2. In Mode 1, the robot moves straight toward the goal. In Mode 2, the robot moves in the tangential direction relative to the gradient of the potential field generated by a stuck robot, choosing the direction closer to the goal out of the two possible options.
    }
    \label{fig:robot_algo}
\end{figure}

Fig.~\ref{fig:robot_algo} shows an overview of the robot's operation. 
The robot dynamically switches between two movement modes: mode 1 and mode 2.
In mode 1, the robot moves directly in the direction of the goal, while in mode 2, it moves along the contour lines of the surrounding potential field. 
In mode 2, the robot adjusts its movement to align with the fixed threshold $g_{th}$ of the measured potential strength, moving perpendicular to the potential gradient. 
By setting this threshold below a certain value, the robot maintains a distance from other robots that at least equals the robot's radius, thereby avoiding collisions. 
This distance is referred to as the circumferential radius.
To simplify the simulation implementation, we assume that in mode 2, the robot moves directly perpendicular to the gradient. 
Additionally, there are two possible perpendicular directions to the gradient: clockwise and counterclockwise. 
In this study, the robot is always configured to choose the direction with the smaller angle relative to the goal direction. 
Hence, the robot's behavior consists of three types of movements: moving straight, turning clockwise, and turning counterclockwise.
When the angle difference between the goal direction and the robot's current direction is less than or equal to $\pi/36\text{rad}$, the robot moves straight at a speed of $1\text{/s}$. 
Otherwise, the robot turns in the direction that reduces the angle with a turning speed of $\pi/6\text{rad/s}$. 

The condition for switching from mode 1 to mode 2 is when the strength of the potential field at the current position of the moving robot $g$ reaches the threshold $g_{th}$, and the angle difference $\theta_{fg}$ between the goal direction and potential field gradient direction satisfies $\theta_{fg} \leq \pi/2\textrm{rad}$. 
This condition indicates that the mobile robot avoids the stuck robot when it has gauged that the distance to the stuck robot $d$ is less than or equal to the circumferential radius $\epsilon$, and it is in the direction of the goal.
On the other hand, the condition for switching from mode 2 to mode 1 is when the angle difference $\theta_{rg}$ between the robot's direction and goal direction satisfies $\theta_{rg}<\pi/2\textrm{rad}$ and the angle difference $\theta_{fg}$ between the potential field gradient direction and goal direction satisfies $\theta_{fg} > \pi/2\textrm{rad}$. 
This is to ensure that the moving robot avoids the potential field and heads towards the goal.

\begin{figure}[t]
    \centering 
     \begin{minipage}[t]{\columnwidth}
      \includegraphics[width=\columnwidth]{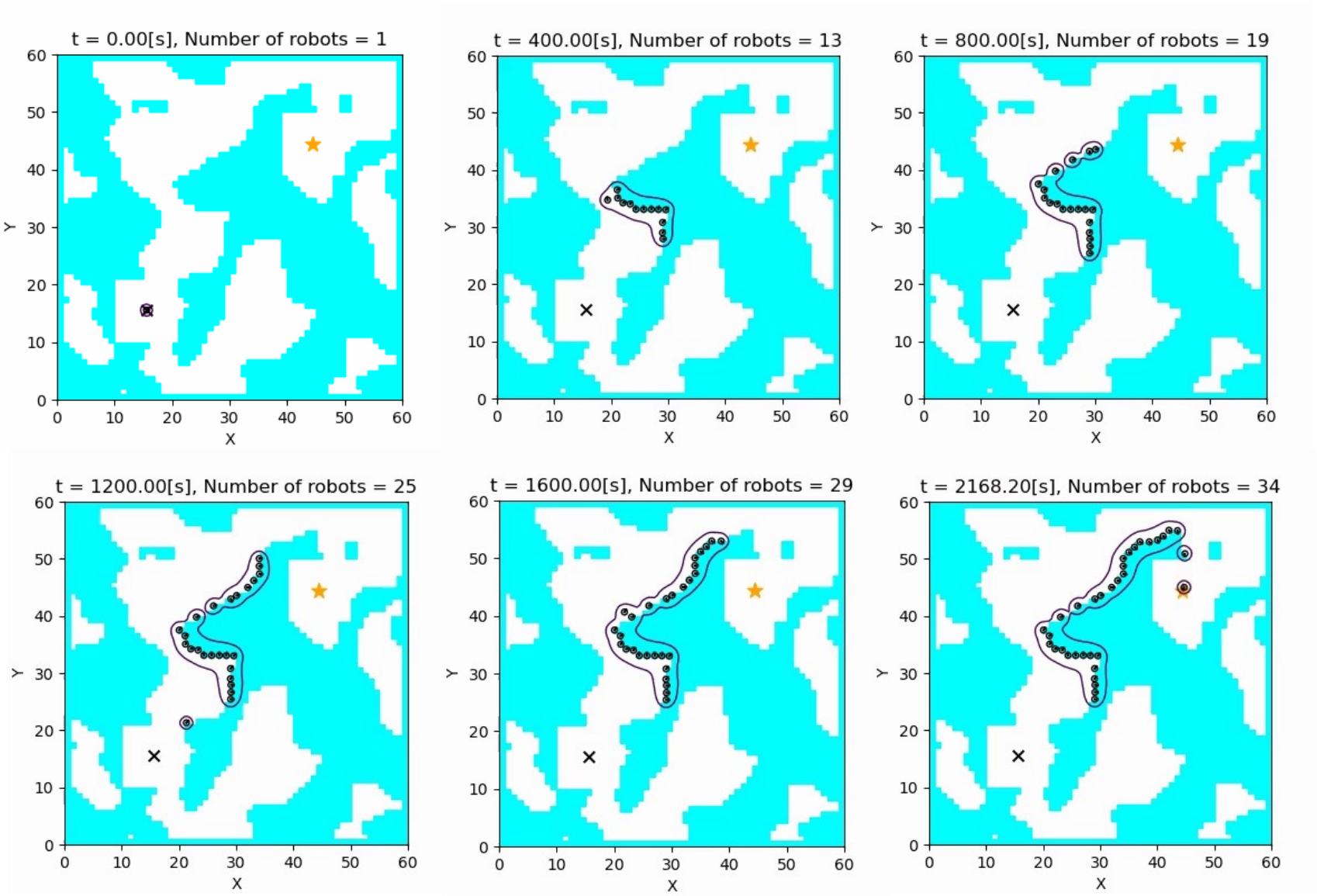}
      \subcaption{Snapshots of successful navigation of an unknown environment}
      \label{fig:sim_sample}
     \end{minipage}
     \begin{minipage}[t]{\columnwidth}
      \includegraphics[width=\columnwidth]{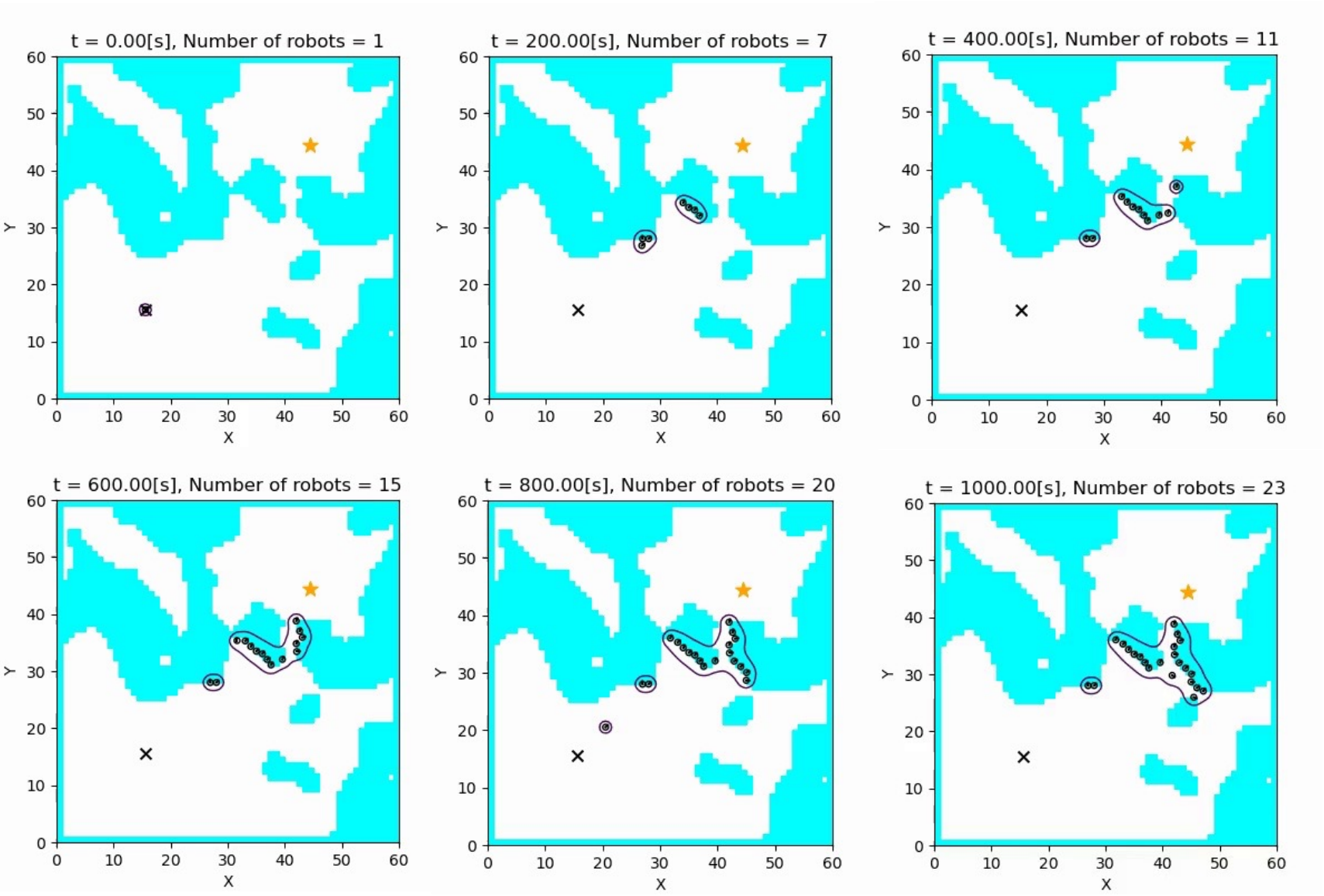}
      \subcaption{Snapshots of unsuccessful navigation of an unknown environment}
      \label{fig:sim_fail}
     \end{minipage}
     \caption{Results of unknown environment navigation based on the proposed method}
    \label{fig:sim_result}
\end{figure}

\subsection{Verification}

An example of the results is shown in Fig.~\ref{fig:sim_result}. 
In this simulation, the threshold for the robot's potential strength is $g_{th} = 1$ and the minimum path width is $3$.
From Fig.~\ref{fig:sim_sample}, it is evident that multiple robots getting stuck creates impassable zones represented by the potential field, allowing subsequent robots to reach the goal. 
This demonstrates that there are conditions under which the robots can successfully navigate unknown environments using potential fields. 
The movie of this simulation can be viewed in Movie1.mp4 in the Supplemental Materials.
However, as shown in Fig.~\ref{fig:sim_fail}, there are also instances where the robots fail to reach the goal. 
The threshold for the robot's potential strength is $g_{th} = 1$ and the minimum path width is $2$.
Therefore, the robots may not be capable of handling all terrains under the given conditions. 
Multiple cases where navigation failed occurred when the narrow paths were blocked by the potential field created by the robot swarm itself. 
Specifically, the successful arrival of the robot at the goal is significantly influenced by the circumferential radius. 
Therefore, in Section~\ref{sec:radius_and_p}, we examine how the circumferential radius and terrain affect the probability of successful navigation.
These simulation videos can be viewed in the Supplemental Materials (Movie1.mp4 and Movie2.mp4).

\section{Simulation analysis}
\label{sec:sim_result}
In this section, we investigate the effects of the minimum path width of the movable area from the start point to the goal and the radius of the circumference of the potential field generated by the robot on the guidance achievement rate.
We also perform simulation validation of the robustness of the guidance to noise in the estimated gradient direction of the potential field.

\subsection{Effect of circumferential radius and minimum path width on navigation performance}
\label{sec:radius_and_p}

As described in Section \ref{sec:Potential}, the robot's circumferential radius $\epsilon$ is adjusted by a constant threshold $g_{th}$ of potential strength. 
By setting the potential field strength threshold $g_{th}$ to $1$, $1/1.5^2$, $1/2^2$, $1/2.5^2$, $1/3^2$, $1/3.5^2$, $1/4^2$, $1/4.5^2$, and $1/5^2$, the circumferential radius $\epsilon$ was varied to be 1, 1.5, 2, 2.5, 3, 3.5, 4, 4.5, and 5 times the grid width for simulations. 
Additionally, for each terrain with minimum path widths of 1, 1.5, 2, 2.5, 3, 3.5, 4, 4.5, and 5 times the grid width, 100 different terrains were generated. 
The percentage of terrains where the robots successfully reached the goal was determined for each circumferential radius.

\begin{figure}[tb]
    \centering 
     \includegraphics[width=\columnwidth]{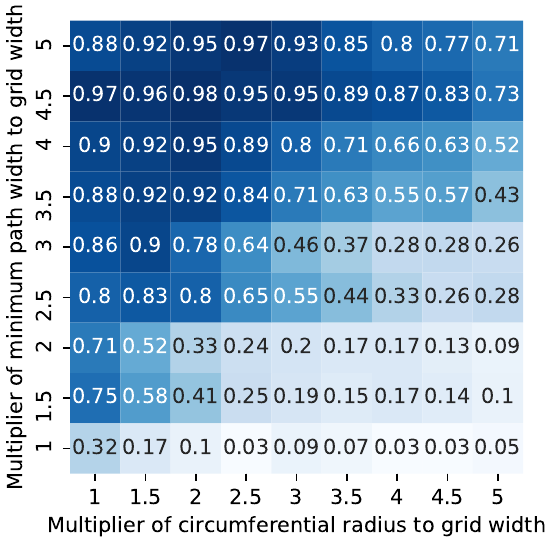}
     \caption{Simulation results on the effect of the circumferential radius and minimum path width on the probability of arrival}
     \label{fig:result_width}
\end{figure}

The simulation results are shown in Fig.~\ref{fig:result_width}. 
The horizontal axis represents the multiplier of the circumferential radius relative to the grid width, and the vertical axis represents the multiplier of the minimum path width relative to the grid width. 
Each cell indicates the success rate. 
Fig.~\ref{fig:result_width} shows that the probability of reaching the goal is larger when the minimum path width (value on the vertical axis) is larger than the robot's radius of circumference (value on the horizontal axis), while the probability is smaller when the minimum path width is smaller than the robot's radius of circumference.
Specifically, we determine that the probability of reaching the goal tends to depend on the relationship between the value of the minimum path width and the value of the radius of circumference of the robot.
This follows from the fact that when the radius of the circumference of the robot is larger than the minimum path width, the potential field generated by the stuck robot blocks the path to the goal.
As described in Section~\ref{sec:Theorem}, the robot can be guaranteed to reach the goal by setting the robot's radius of circumference $\epsilon$ to a value smaller than the minimum path width $l$ of the path to the goal in order to achieve successful navigation.
Thus, we have demonstrated the efficacy of the implementation of BYCOMS with potential fields.

\subsection{Robustness to local sensing noise}
\label{sec:Robust}
\begin{figure}[t]
    \centering 
     \includegraphics[width=\columnwidth]{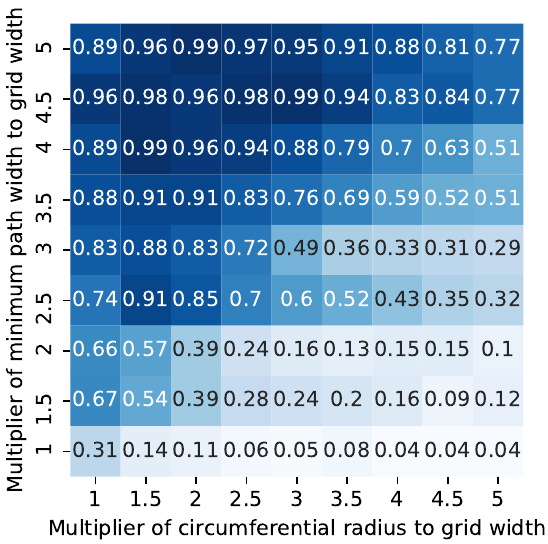}
     \caption{Simulation results with noise in the observed values for the effect of the circumference radius and minimum path width on the probability of arrival}
     \label{fig:result_width_noise}
\end{figure}

We examined the behavior of the proposed system when noise was added to the measurements of the potential field's strength and gradient direction. 
In order to introduce noise in the potential field's strength, a uniformly random value in the range of $0.8$ to $1.2$ was multiplied by the measured value to obtain the new measurement. 
To introduce noise in the gradient direction of the potential field, a uniformly random value in the range of $-\pi/6$ to $\pi/6\textrm{rad}$ was added to the gradient direction to obtain the new direction. 
The terrain used was the same as that described in Section~\ref{sec:radius_and_p}. 
The reachability probabilities considering the sensing noise are shown in Fig.~\ref{fig:result_width_noise}. 
In particular, we can observe that the probability of reaching the goal depends on the relationship between the radius of circumference of the robot and the minimum path width of the goal path, as well as the result without considering noise (Fig.~\ref{fig:result_width}).
Compared to the results shown in Fig.~\ref{fig:result_width}, the probability of reaching the goal does not change significantly under most conditions.
The increase in the probability of reaching the goal with noise can be attributed to the randomness of the noise having a positive effect on the probability of reaching the goal by chance.
Thus, the simulation results demonstrate the robustness of BYCOMS against sensing noise.

\section{Experimental verification of BYCOMS based on acoustic fields}
\label{sec:Expe}

In this section, we conduct navigation experiments with a swarm of robots to verify whether BYCOMS has the ability to traverse unknown environments in the real world. 
We developed robots that express the potential field through the intensity of sound, generate acoustic fields, and sense and move along the gradients of these fields.

\subsection{Development of acoustic field based robot}
\label{sec:Robo}

\begin{figure}[t]
    \centering 
     \begin{minipage}[t]{\columnwidth}
      \includegraphics[width=\columnwidth]{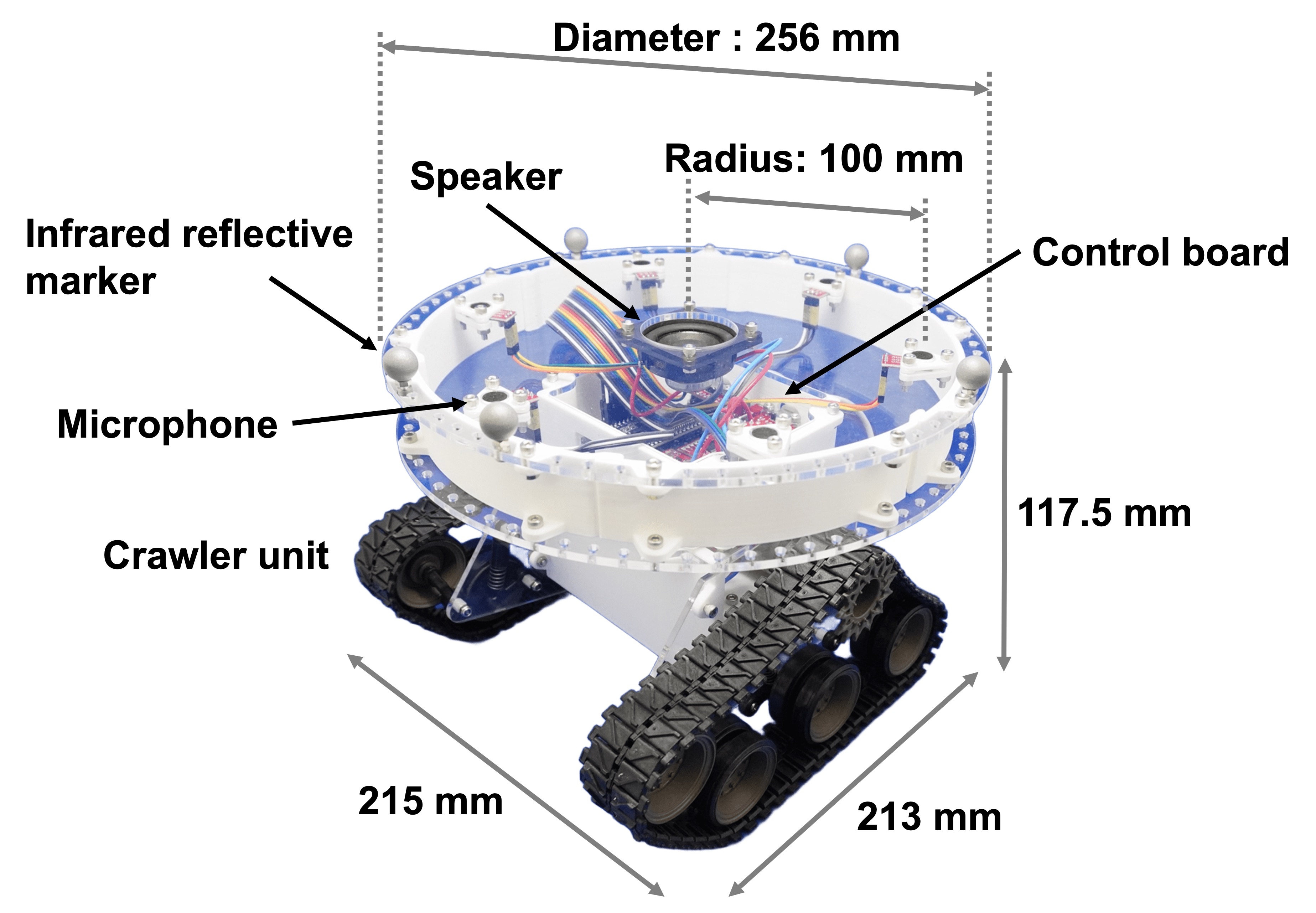}
      \subcaption{Diagonal view}
      \label{fig:diagonal}
     \end{minipage}
     \begin{minipage}[t]{\columnwidth}
      \includegraphics[width=\columnwidth]{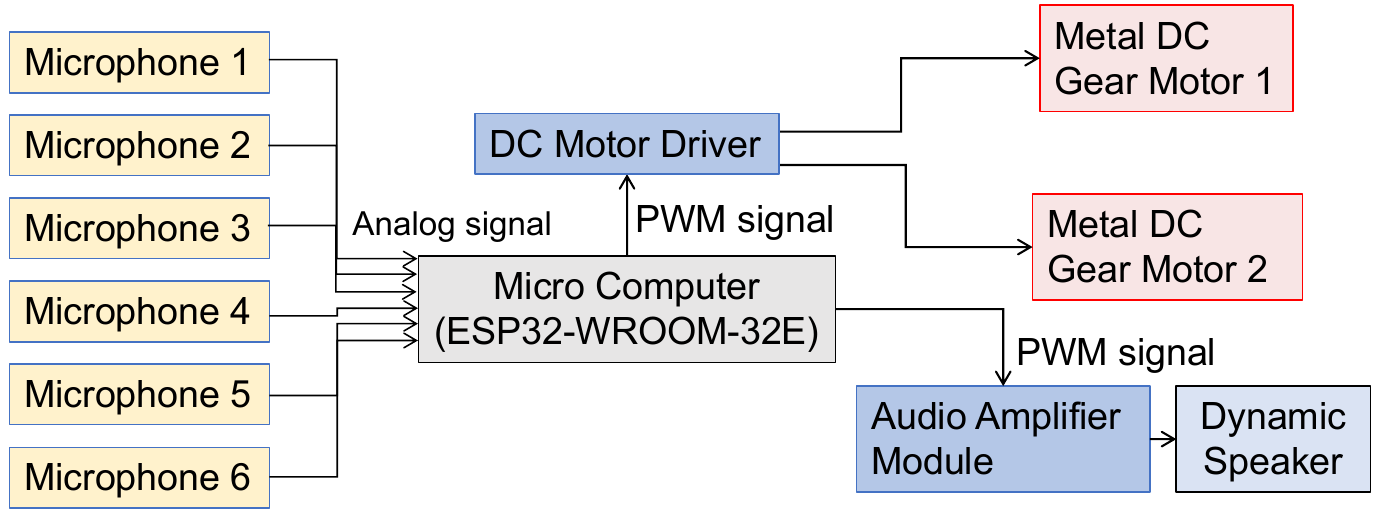}
      \subcaption{Control system}
      \label{fig:control_system}
     \end{minipage}
    \caption{Overview of the developed robot}
    \label{fig:robot_overview}
\end{figure}

Fig.~\ref{fig:robot_overview} shows an overview of the developed robot. 
The robot consists of an upper microphone array for sound field gradient sensing and a lower crawler unit for movement. 
The dimensions of the robot are as follows: $213\textrm{mm}$ in length, $215\textrm{mm}$ in width, and $117.5\textrm{mm}$ in height. 
The microphone array is unitized in an acrylic plate case with a diameter of $256\textrm{mm}$, centered around the middle of the robot. 
The weight of the robot is $1.78\textrm{kg}$.
The robot is controlled by an ESP32-WROOM-32E (Espressif Systems) microcontroller. 
In addition, it is equipped with motors and a crawler unit (Devastator Tank, DFROBOT) for movement, a motor driver (ROB-14450, SparkFun), and a mobile battery (DE-C17L-5000, Elecom). 
To generate the sound field, the robot has a speaker (S0266, DAYTON AUDIO) connected to a sound signal amplifier module (BOB-11044, SparkFun) positioned at the center on the top of the robot. 
For gradient measurement, it is equipped with six condenser microphones (BOB-12758, SparkFun) placed at equal intervals on a circle with a radius of $100\textrm{mm}$. 
The structural components of the robot are made of $3\textrm{mm}$ laser-cut acrylic sheets and 3D printed PLA parts. 
Additionally, the robot has five infrared reflective markers on the top, which allow an optical motion capture system (Optitrack) to log the robot's position and orientation.

\subsection{Measuring the gradient direction of the sound field}
\label{sec:calc_grad}
In the proposed method using acoustic fields, robots measure the gradient direction of the acoustic field at their current location.
In this experiment, we adopt a method of measuring the gradient direction by placing multiple microphones on a circumference and using the differences in their measured values.
Consider a case where 6 microphones $M_1, M_2, \cdots ,M_6$ are evenly spaced on a circle of radius $r$ that shares its center with the robot.
We employ $xy$ coordinates where the center of the robot is considered the origin, and $M_1$ is at $(r,0)$.
Let the measured sound intensities at each microphone be $V_1, V_2, \cdots ,V_6$.
The sound intensity at the center of the robot is estimated as the average of all microphone measurements, expressed by the following equation:

\begin{equation}
    V_{ave} = \sum_{n=1}^{6} \frac{V_n}{6}.
\end{equation}
Moreover, let $\bm{m_n}$ be the position vector of the $n$-th microphone relative to the center of the robot.
The estimated gradient of the acoustic field obtained by the robot is expressed by the following equation:
\begin{equation}
    \bm{grad} = \sum_{n=1}^{6} (V_n-V_{ave}) \bm{m_n}.
\end{equation}
The robot normalizes this $\bm{grad}$ and uses only the direction of the acoustic field gradient as the estimate.

\subsection{Robot algorithm}
\begin{figure}[t]
    \centering 
     \includegraphics[width=\columnwidth]{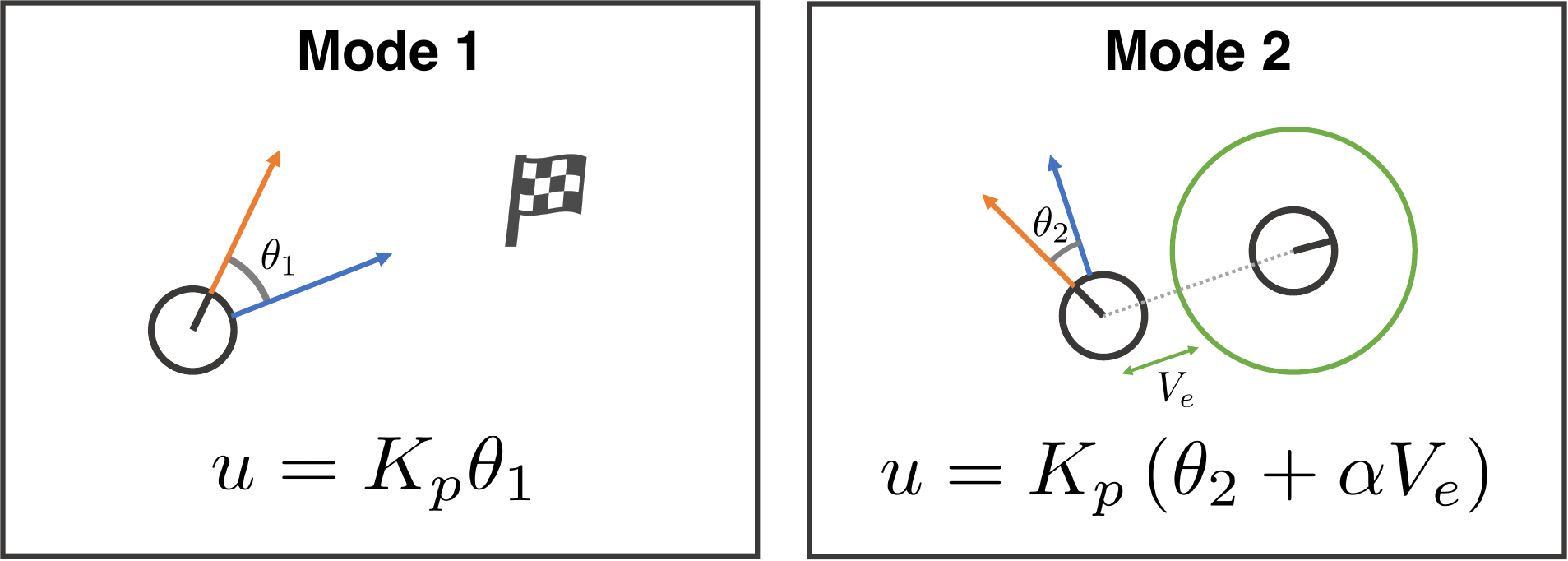}
     \caption{Robot control methods in experiments}
     \label{fig:robot_control}
\end{figure}

The overview of the robot control method is shown in Fig.~\ref{fig:robot_control}.
As described in Section~\ref{sec:Sim}, the robot is implemented with mode 1, which moves directly to the goal, and mode 2, which moves along the sound field.
In mode 1, the robot uses the angle difference $\theta_1 ( -\pi < \theta_1 \leq \pi)$ between the direction of the goal and its own direction, and the control input is $u = K_p \theta_1$.
The robot was controlled by applying a voltage of $v_m + u$ to the left motor and $v_m - u$ to the right motor, with $v_m$ as the reference value of the voltage applied to the motors.
In mode 2, the robot measures the gradient direction of the sound field and sets the target direction as the direction rotated by $\pi/2$ from the gradient. 
It calculates the angle difference $\theta_2 ( -\pi < \theta_2 \leq \pi)$ between this target direction and its own direction. 
Additionally, the target value of the magnitude of the microphone measurements is set as $V_{target}$, and the difference between $V_{target}$ and average of all microphone measurements $V_{ave}$ is calculated as $V_e = V_{ave} - V_{target}$. 
Using these values, the control input is $u = K_p (\theta_2 + \alpha V_e)$. $\alpha$ is the weighting coefficient for $\theta_2$ and $V_e$.
As in mode 1, the robot was controlled by applying a voltage of $v_m + u$ to the left motor and $v_m - u$ to the right motor.
The switching conditions between the two modes are as follows. 
The condition for switching from mode 1 to mode 2 is $V_{ave} > V_{target}$ and the angle between the direction of the goal and $\bm{grad}$ is less than $\pi/2$. 
The condition for switching from mode 2 to mode 1 is the angle between the direction of the robot and the goal is less than $\pi/2$ while that between $\bm{grad}$ and the goal direction is greater than $\pi/2$, or when $V_{ave}$ significantly decreases below the lower threshold value $V_{min}$. 
The latter is set to account for situations where $\bm{grad}$ cannot be measured because the robot has moved too far from the sound source in mode 2.
The parameters for this experiment are $K_p = 100$, $\alpha = 0.0001$, $V_{target} = 22000$, and $V_{min} = 15000$.

In our proposed navigation method, each robot must emit sound while also listening for sound  signals from other robots. 
Theoretically, if all the robots emit the same frequency, the robot's own sound would be heard equally by all its microphones, thus not affecting the calculation of its gradient direction. 
However, in practical scenarios, measurement errors might cause the robot's own sound to interfere with gradient measurement. 
Additionally, vibrations transmitted through the robot's structural components could introduce significant noise.
Therefore, we vary the sound frequency for each robot in our experiments. 
Each robot can isolate the sounds emitted by others using fast fourier transform (FFT) \cite{nussbaumer1982fast}. 
The robot extracts the sound of other robots' frequencies from the measured sound and uses the sum of their amplitude spectra as the microphone measurements. 
This processed data is then used to calculate the gradient as described in Section \ref{sec:calc_grad}.

\begin{figure}[t]
    \centering 
     \includegraphics[width=0.9\columnwidth]{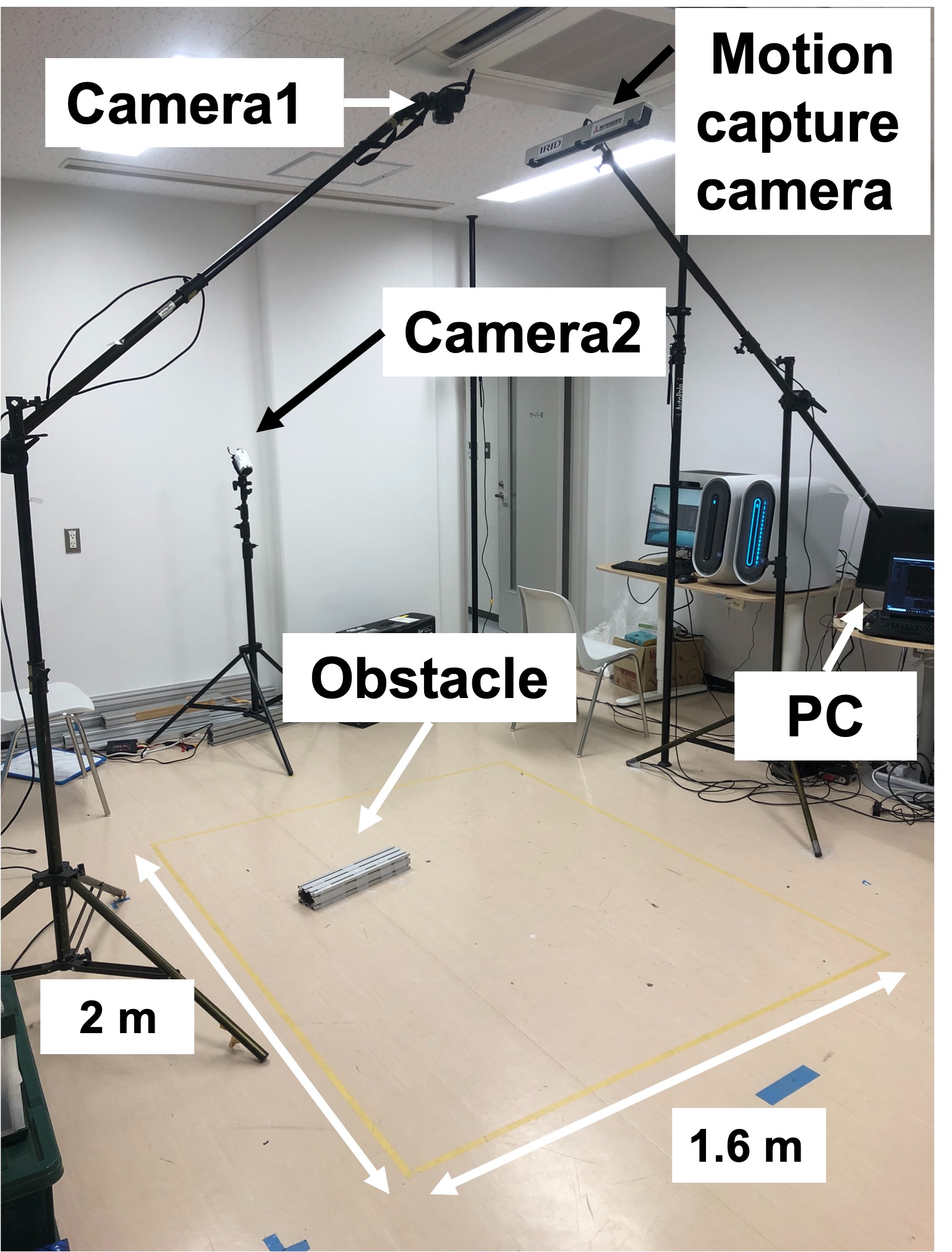}
     \caption{Experimental enviroment}
     \label{fig:exp_env}
\end{figure}
\begin{figure}[t]
    \centering 
     \includegraphics[width=\columnwidth]{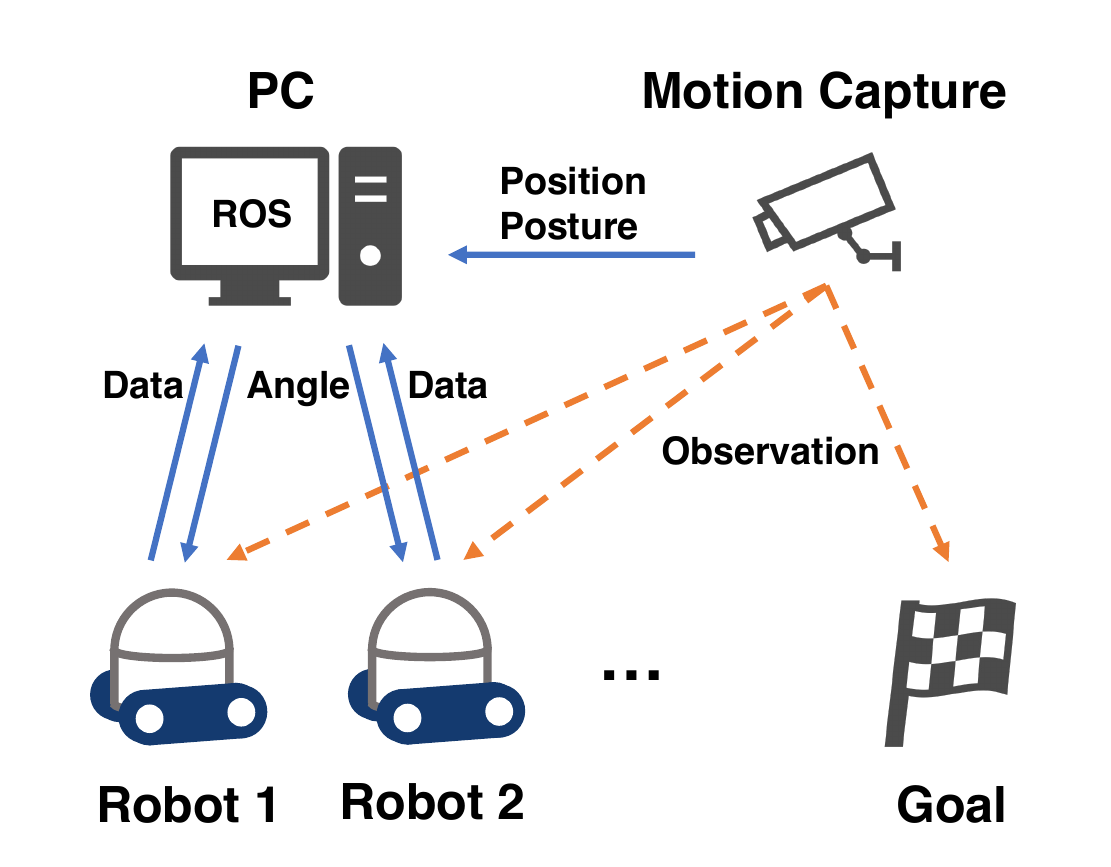}
     \caption{Schematic diagram of the system employed in the experiments}
     \label{fig:robot_system}
\end{figure}

\subsection{Experimental environment}
\label{sec:Result}
We validated the effectiveness of the proposed method in real-world environments through experimental verification with three developed robots. 
The experimental environment is shown in Fig.~\ref{fig:exp_env}. 
An acrylic board with reflective markers was placed at the goal. 
To emulate an unknown environment for the robots, aluminum frames were set up as obstacles on the way to the goal. 
The later robot starts moving from the starting point after the earlier robot gets stuck on an obstacle.
Additionally, each of the three robots emitted sine wave sounds from their speakers at frequencies of $224 \mathrm{Hz}$, $320 \mathrm{Hz}$, and $512 \mathrm{Hz}$, respectively.

The overview of the system controlling the robot swarm is shown in Fig.~\ref{fig:robot_system}. 
First, the position and orientation of the robots and the position of the goal are measured using the motion capture system and sent to the main PC. 
On the PC, all the data are processed as topics using Robot Operating System (ROS) \cite{quigley2009ros}. 
Based on the sensor data obtained by the motion capture system, the direction of the goal from each robot's perspective is calculated and sent to each robot via WiFi. 
The robots then move based on the transmitted goal direction data, and the locally sensed sound field gradient. 
Additionally, the measured sound field gradient data is sent to the main PC, where the data is stored.

\subsection{Experimental results}
\label{sec:Okunai}
\begin{figure}[t]
    \centering 
     \begin{minipage}[t]{\columnwidth}
      \includegraphics[width=\columnwidth]{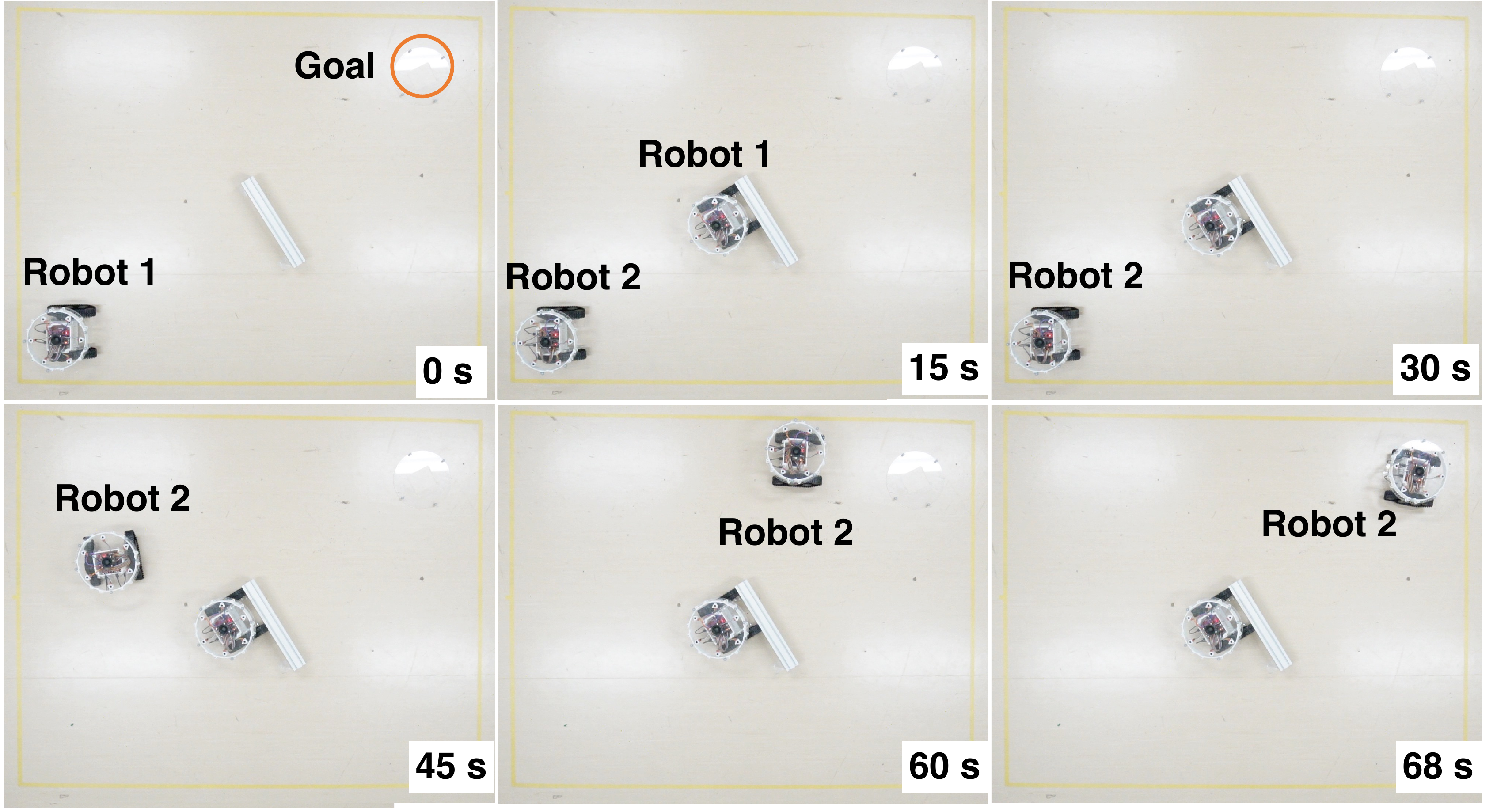}
      \subcaption{Snapshots of the behavior of two robots in the navigation}
      \label{fig:exp_2robots}
     \end{minipage}
     \begin{minipage}[t]{\columnwidth}
      \includegraphics[width=\columnwidth]{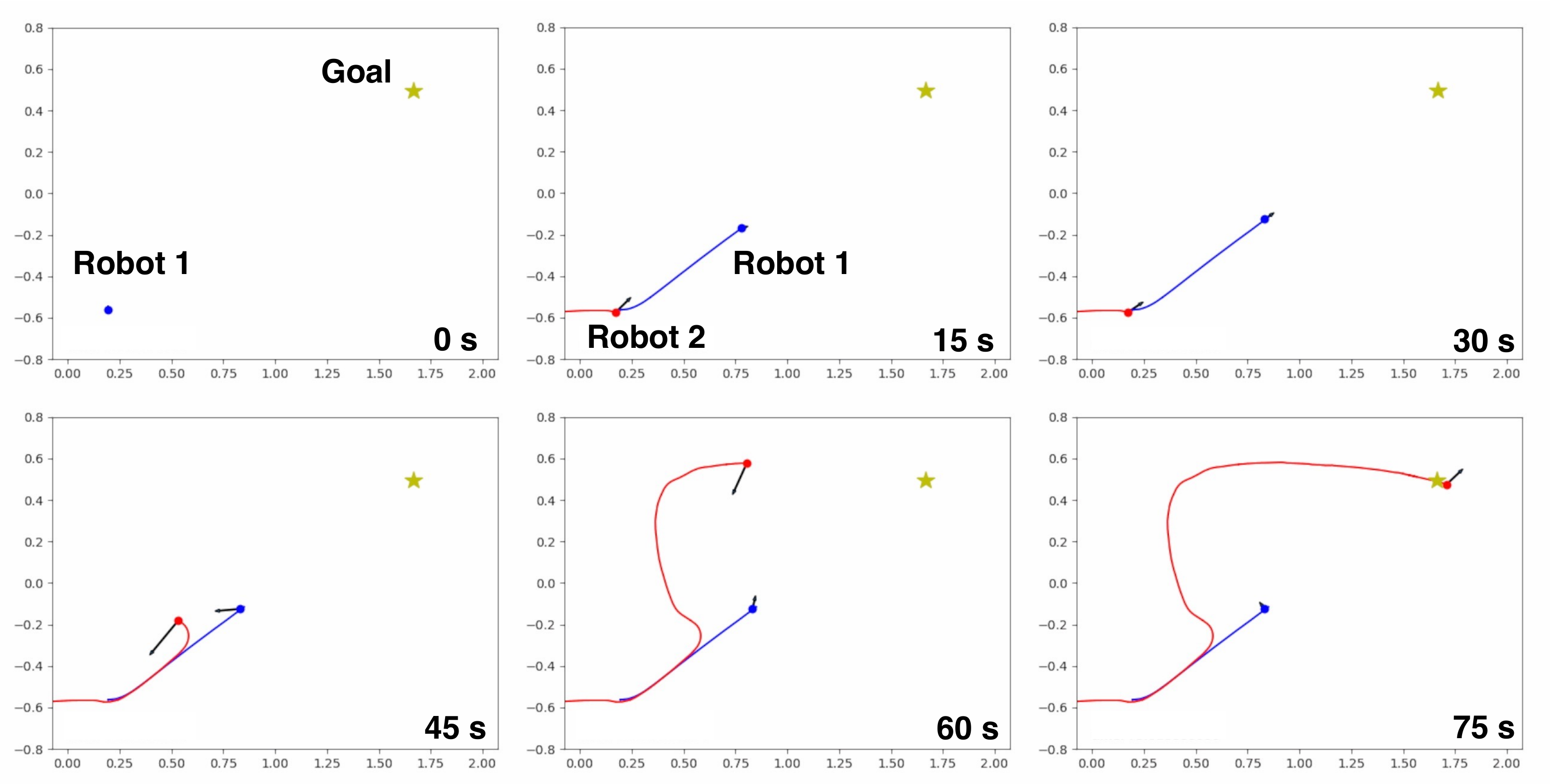}
      \subcaption{Snapshots of the measurement data of two robots}
      \label{fig:bag_2robots}
     \end{minipage}
    \caption{Experimental results with two robots}
    \label{fig:exp_inside_short}
\end{figure}

\begin{figure}[t]
    \centering 
     \begin{minipage}[t]{\columnwidth}
      \includegraphics[width=\columnwidth]{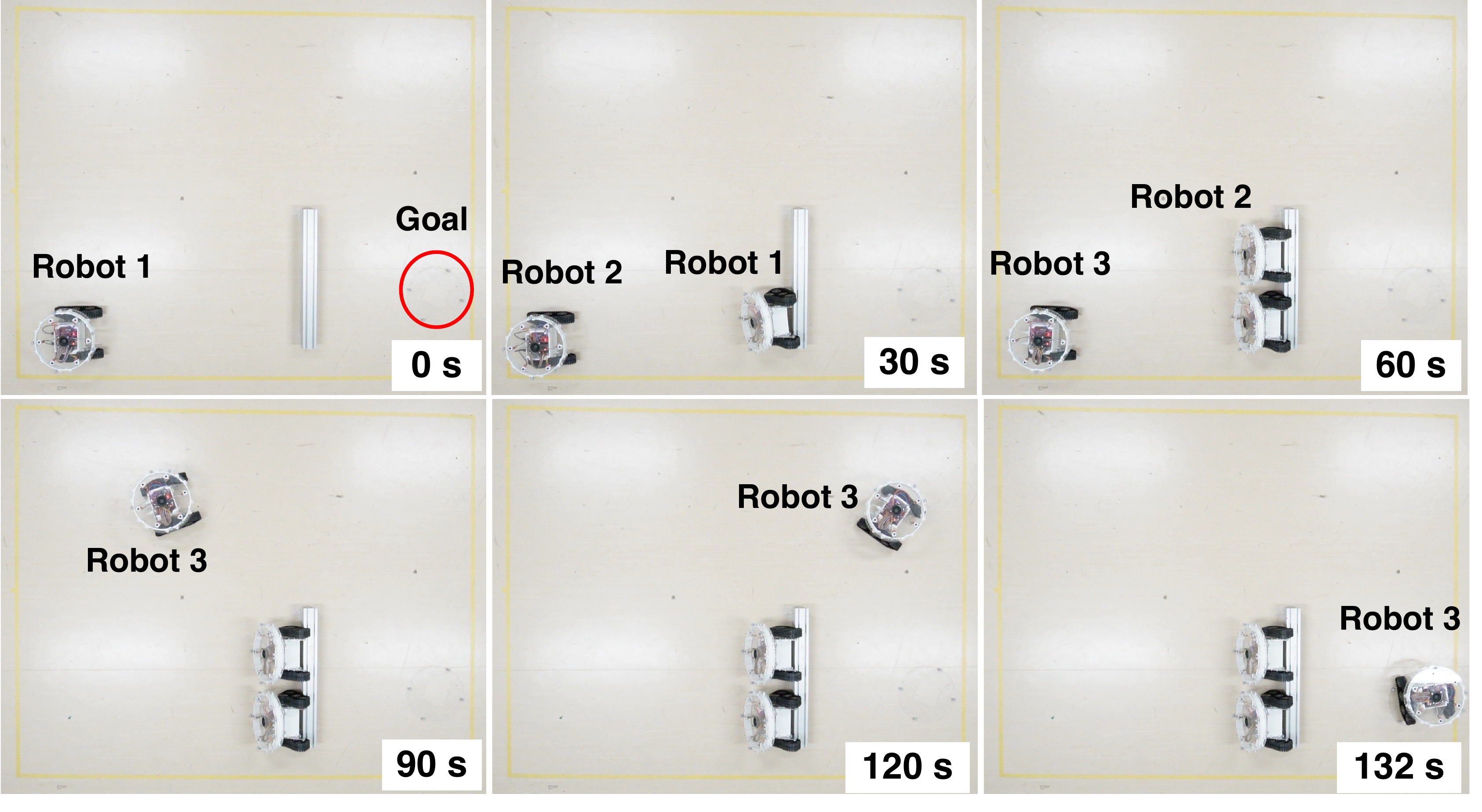}
      \subcaption{Snapshots of the behavior of three robots in the navigation}
      \label{fig:exp_3robots}
     \end{minipage}
     \begin{minipage}[t]{\columnwidth}
      \includegraphics[width=\columnwidth]{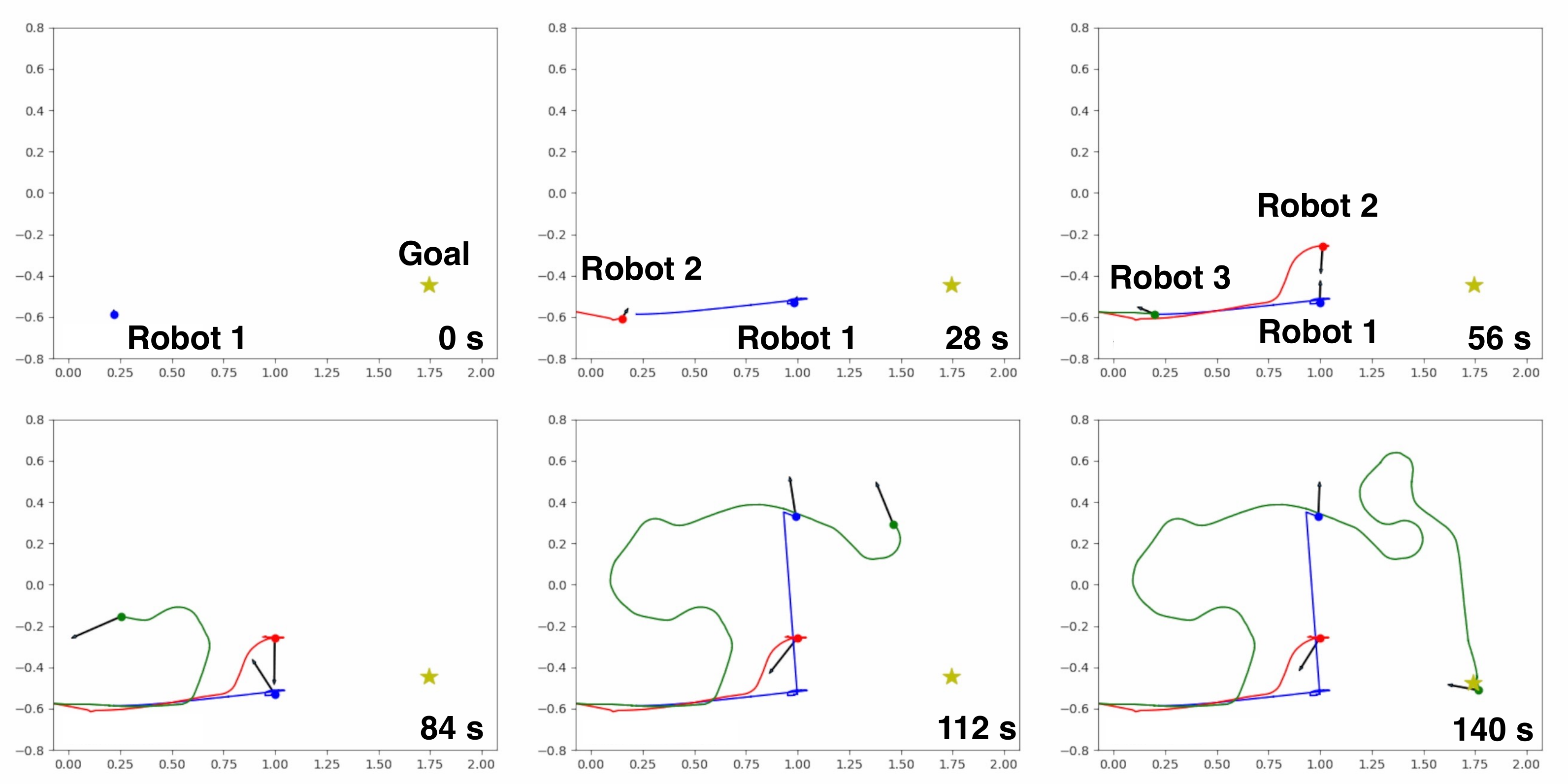}
      \subcaption{Snapshots of the measurement data of three robots}
      \label{fig:bag_3robots}
     \end{minipage}
    \caption{Experimental results with three robots}
    \label{fig:exp_inside_long}
\end{figure}

In this study, we conducted two experiments in which the goal was obstructed by two types of obstacles of different lengths.  
The experimental results are shown in Fig.~\ref{fig:exp_inside_short} and Fig.~\ref{fig:exp_inside_long}.  
In Fig.~\ref{fig:exp_2robots}, where a short obstacle was installed, it can be observed that the first robot got stuck on the aluminum frame, while the second robot avoided both the sound field of the first robot and the obstacle, thus reaching the goal.  
In Fig.~\ref{fig:exp_3robots}, with a long obstacle installed, the second robot similarly bypassed the first robot, but it also got stuck owing to the presence of an obstacle in its path.  
However, the third robot, guided by the combined sound fields of the first and second robots, successfully avoided the obstacle and reached the goal.  

Fig.~\ref{fig:bag_2robots} and Fig.~\ref{fig:bag_3robots} show the positions of the robots and the goal, measured gradient directions, and microphone measurements.  
Blue dots represent Robot 1, red dots represent Robot 2, green dots represent Robot 3, stars indicate the goal, arrow directions show the measured gradient directions, and arrow lengths represent the average microphone measurements.  
However, in Fig.~\ref{fig:bag_3robots}, after Robot 1 got stuck, its position coordinates were mistakenly identified as those of Robot 3.  
This was owing to the rigid body definition based on the detachment of multiple markers on the robot.  
From Fig.~\ref{fig:bag_2robots} and Fig.~\ref{fig:bag_3robots}, it can be observed that when robots are close to each other, they can measure the direction of each others sound fields.  
However, when the robots are slightly apart, they do not measure as well.  
This is probably because the difference in microphone measurements becomes small, and the influence of noise, such as echoes and motor vibrations becomes significant.  
Consequently, it can be considered that in this experiment, the P control based on the average microphone measurement has a greater influence on the robot movement than the gradient direction.  
Therefore, the robots based on the sound field were shown to have the capability to traverse unknown environments even in real-world experiments.
These experiments videos can be viewed in the Supplemental Materials (Movie3.mp4 and Movie4.mp4).

\section{Conclusion}
\label{sec:Conc}
This study proposed a minimal navigation algorithm for a swarm robot system to traverse unknown environments. 
We mathematically demonstrated that the entire swarm could traverse unknown environments by  enabling each robot to generate and follow a path to the goal while bypassing other robots in real-time. 
We implemented the proposed unknown environment navigation method based on the modification of potential fields in simulations. 
The simulation analysis results showed that a smaller turning radius increases the success rate, and it is robust to local sensing noise. 
In addition, we developed a swarm of robots that generate and sense a sound field using speakers and microphone arrays, estimate the sound field gradient, and perform avoidance maneuvers relative to surrounding robots. 
The results of the swarm robot navigation experiments demonstrated that the proposed method is feasible in real-world scenarios. 
A noteworthy aspect of this study is that it demonstrated the potential for the entire system to adaptively traverse unknown environments by leveraging the principle of strength in numbers, even if each robot does not possess high sensing capabilities, computational power, or environmental traversal performance. 
This proposed method may serve as a standard for swarm robot navigation in various environments. 
However, it is also true that swarm robot navigation in the real world is not straightforward. 
This study assumes an infinite number of robots, which is unrealistic owing to development costs and time constraints. 
Additionally, it assumes the constant existence of a viable path from the starting point to the goal, which may not always be the case. 
These issues are closely related to how much we enhance the performance of individual robots. 
By improving the software and hardware functions of each robot, we can reduce the number of robots sacrificed;thus, reducing the necessary number of robots.
Future research will clarify the relationship between the number of robots needed to achieve system-wide navigation and the functions provided to each robot, based on the proposed method. 
Moreover, we plan to develop methods to achieve more efficient goal attainment by temporally varying the information shared among robots in BYCOMS and introducing individual differences among the robots.

\section*{Acknowledgments}
This research was supported in part by grants-in-aid for JSPS KAKENHI Grant Number JP22K14277, JP24H01439, and JST Moonshot Research and Development Program JPMJMS2032 (Innovation in Construction of Infrastructure with Cooperative AI and Multi Robots Adapting to Various Environments).

{\small
\bibliography{myref}

\begin{thebibliography}{10}
\providecommand{\url}[1]{#1}
\csname url@samestyle\endcsname
\providecommand{\newblock}{\relax}
\providecommand{\bibinfo}[2]{#2}
\providecommand{\BIBentrySTDinterwordspacing}{\spaceskip=0pt\relax}
\providecommand{\BIBentryALTinterwordstretchfactor}{4}
\providecommand{\BIBentryALTinterwordspacing}{\spaceskip=\fontdimen2\font plus
\BIBentryALTinterwordstretchfactor\fontdimen3\font minus
  \fontdimen4\font\relax}
\providecommand{\BIBforeignlanguage}[2]{{%
\expandafter\ifx\csname l@#1\endcsname\relax
\typeout{** WARNING: IEEEtran.bst: No hyphenation pattern has been}%
\typeout{** loaded for the language `#1'. Using the pattern for}%
\typeout{** the default language instead.}%
\else
\language=\csname l@#1\endcsname
\fi
#2}}
\providecommand{\BIBdecl}{\relax}
\BIBdecl

\bibitem{open}
\BIBentryALTinterwordspacing
K.~Nagatani, M.~Abe, K.~Osuka, P.~jo~Chun, T.~Okatani, M.~Nishio, S.~Chikushi,
  T.~Matsubara, Y.~Ikemoto, and H.~Asama, ``Innovative technologies for
  infrastructure construction and maintenance through collaborative robots
  based on an open design approach,'' \emph{Advanced Robotics}, vol.~35, pp.
  715--722, 2021. [Online]. Available:
  \url{https://www.tandfonline.com/doi/abs/10.1080/01691864.2021.1929471}
\BIBentrySTDinterwordspacing

\bibitem{thrun2005multi}
S.~Thrun and Y.~Liu, ``Multi-robot slam with sparse extended information
  filers,'' in \emph{Robotics Research. The Eleventh International Symposium:
  With 303 Figures}.\hskip 1em plus 0.5em minus 0.4em\relax Springer, 2005, pp.
  254--266.

\bibitem{ng2007performance}
J.~Ng and T.~Br{\"a}unl, ``Performance comparison of bug navigation
  algorithms,'' \emph{Journal of Intelligent and Robotic Systems}, vol.~50, pp.
  73--84, 2007.

\bibitem{malavazi2018lidar}
F.~B. Malavazi, R.~Guyonneau, J.-B. Fasquel, S.~Lagrange, and F.~Mercier,
  ``Lidar-only based navigation algorithm for an autonomous agricultural
  robot,'' \emph{Computers and electronics in agriculture}, vol. 154, pp.
  71--79, 2018.

\bibitem{1638022}
H.~Durrant-Whyte and T.~Bailey, ``Simultaneous localization and mapping: part
  i,'' \emph{IEEE Robotics \& Automation Magazine}, vol.~13, no.~2, pp.
  99--110, 2006.

\bibitem{lumelsky1987path}
V.~J. Lumelsky and A.~A. Stepanov, ``Path-planning strategies for a point
  mobile automaton moving amidst unknown obstacles of arbitrary shape,''
  \emph{Algorithmica}, vol.~2, no.~1, pp. 403--430, 1987.

\bibitem{4209200}
S.~Sarid, A.~Shapiro, and Y.~Gabriely, ``Mrbug: A competitive multi-robot path
  finding algorithm,'' in \emph{Proceedings 2007 IEEE International Conference
  on Robotics and Automation}, 2007, pp. 877--882.

\bibitem{atanasov2015decentralized}
N.~Atanasov, J.~Le~Ny, K.~Daniilidis, and G.~J. Pappas, ``Decentralized active
  information acquisition: Theory and application to multi-robot slam,'' in
  \emph{2015 IEEE International Conference on Robotics and Automation
  (ICRA)}.\hskip 1em plus 0.5em minus 0.4em\relax IEEE, 2015, pp. 4775--4782.

\bibitem{10.3389/frobt.2021.618268}
\BIBentryALTinterwordspacing
M.~Kegeleirs, G.~Grisetti, and M.~Birattari, ``Swarm slam: Challenges and
  perspectives,'' \emph{Frontiers in Robotics and AI}, vol.~8, 2021. [Online].
  Available:
  \url{https://www.frontiersin.org/journals/robotics-and-ai/articles/10.3389/frobt.2021.618268}
\BIBentrySTDinterwordspacing

\bibitem{kegeleirs2021swarm}
------, ``Swarm slam: Challenges and perspectives,'' \emph{Frontiers in
  Robotics and AI}, vol.~8, p. 618268, 2021.

\bibitem{mcguire2019minimal}
K.~McGuire, C.~De~Wagter, K.~Tuyls, H.~Kappen, and G.~C. de~Croon, ``Minimal
  navigation solution for a swarm of tiny flying robots to explore an unknown
  environment,'' \emph{Science Robotics}, vol.~4, no.~35, p. eaaw9710, 2019.

\bibitem{lee2020learning}
J.~Lee, J.~Hwangbo, L.~Wellhausen, V.~Koltun, and M.~Hutter, ``Learning
  quadrupedal locomotion over challenging terrain,'' \emph{Science robotics},
  vol.~5, no.~47, p. eabc5986, 2020.

\bibitem{doi:10.1126/scirobotics.abk2822}
\BIBentryALTinterwordspacing
T.~Miki, J.~Lee, J.~Hwangbo, L.~Wellhausen, V.~Koltun, and M.~Hutter,
  ``Learning robust perceptive locomotion for quadrupedal robots in the wild,''
  \emph{Science Robotics}, vol.~7, no.~62, p. eabk2822, 2022. [Online].
  Available: \url{https://www.science.org/doi/abs/10.1126/scirobotics.abk2822}
\BIBentrySTDinterwordspacing

\bibitem{zheng2021learning}
J.~Zheng, T.~Zhang, C.~Wang, M.~Xiong, and G.~Xie, ``Learning for attitude
  holding of a robotic fish: An end-to-end approach with sim-to-real
  transfer,'' \emph{IEEE Transactions on Robotics}, vol.~38, no.~2, pp.
  1287--1303, 2021.

\bibitem{10.1007/978-3-540-30217-9_86}
R.~Gro{\ss} and M.~Dorigo, ``Group transport of an object to a target that only
  some group members may sense,'' in \emph{Parallel Problem Solving from Nature
  - PPSN VIII}, X.~Yao, E.~K. Burke, J.~A. Lozano, J.~Smith, J.~J.
  Merelo-Guerv{\'o}s, J.~A. Bullinaria, J.~E. Rowe, P.~Ti{\v{n}}o,
  A.~Kab{\'a}n, and H.-P. Schwefel, Eds.\hskip 1em plus 0.5em minus 0.4em\relax
  Berlin, Heidelberg: Springer Berlin Heidelberg, 2004, pp. 852--861.

\bibitem{1362262944632390912}
\BIBentryALTinterwordspacing
R.~Michael, C.~Alejandro, and N.~Radhika, ``Programmable self-assembly in a
  thousand-robot swarm,'' \emph{Science}, vol. 345, no. 6198, pp. 795--799, 08
  2014. [Online]. Available:
  \url{https://cir.nii.ac.jp/crid/1362262944632390912}
\BIBentrySTDinterwordspacing

\bibitem{doi:10.1177/0037549720930082}
\BIBentryALTinterwordspacing
J.~J. Kandathil, R.~Mathew, and S.~S. Hiremath, ``Development and analysis of a
  novel obstacle avoidance strategy for a multi-robot system inspired by the
  bug-1 algorithm,'' \emph{SIMULATION}, vol.~96, no.~10, pp. 807--824, 2020.
  [Online]. Available: \url{https://doi.org/10.1177/0037549720930082}
\BIBentrySTDinterwordspacing

\bibitem{sugawara2018casualty}
K.~Sugawara, Y.~Doi, and M.~Shishido, ``Casualty-based cooperation in swarm
  robots,'' \emph{Artificial Life and Robotics}, vol.~23, no.~4, pp. 645--650,
  2018.

\bibitem{fujisawa2008communication}
R.~Fujisawa, H.~Imamura, T.~Hashimoto, and F.~Matsuno, ``Communication using
  pheromone field for multiple robots,'' in \emph{2008 IEEE/RSJ International
  Conference on Intelligent Robots and Systems}.\hskip 1em plus 0.5em minus
  0.4em\relax IEEE, 2008, pp. 1391--1396.

\bibitem{sugawara2004foraging}
K.~Sugawara, T.~Kazama, and T.~Watanabe, ``Foraging behavior of interacting
  robots with virtual pheromone,'' in \emph{2004 IEEE/RSJ International
  Conference on Intelligent Robots and Systems (IROS) (IEEE Cat.
  No.04CH37566)}, vol.~3, 2004, pp. 3074--3079 vol.3.

\bibitem{YuichiroSUEOKA202120-00280}
Y.~SUEOKA, D.~D. Khanh, Y.~TSUNODA, Y.~SUGIMOTO, and K.~OSUKA, ``Analysis and
  experiment of robot navigation by sound field using interaction with
  obstacles,'' \emph{Transactions of the JSME}, vol.~87, no. 896, p.
  tsunoda2019experimental, 2021 (in japanese).

\bibitem{tsunoda2019experimental}
Y.~Tsunoda, Y.~Sueoka, and K.~Osuka, ``Experimental analysis of acoustic field
  control-based robot navigation,'' \emph{Journal of Robotics and
  Mechatronics}, vol.~31, no.~1, pp. 110--117, 2019.

\bibitem{Yusuke_Tsunoda2023}
Y.~Tsunoda, L.~T. Nghia, Y.~Sueoka, and K.~Osuka, ``Experimental analysis of
  shepherding-type robot navigation utilizing sound-obstacle-interaction,''
  \emph{Journal of Robotics and Mechatronics}, vol.~35, no.~4, pp. 957--968,
  2023.

\bibitem{rose2016algorithms}
T.~J. Rose and A.~G. Bakaoukas, ``Algorithms and approaches for procedural
  terrain generation-a brief review of current techniques,'' in \emph{2016 8th
  International Conference on Games and Virtual Worlds for Serious Applications
  (VS-GAMES)}.\hskip 1em plus 0.5em minus 0.4em\relax IEEE, 2016, pp. 1--2.

\bibitem{warren1993fast}
C.~W. Warren, ``Fast path planning using modified a* method,'' in \emph{[1993]
  Proceedings IEEE International Conference on Robotics and Automation}.\hskip
  1em plus 0.5em minus 0.4em\relax IEEE, 1993, pp. 662--667.

\bibitem{nussbaumer1982fast}
H.~J. Nussbaumer and H.~J. Nussbaumer, \emph{The fast Fourier transform}.\hskip
  1em plus 0.5em minus 0.4em\relax Springer, 1982.

\bibitem{quigley2009ros}
M.~Quigley, K.~Conley, B.~Gerkey, J.~Faust, T.~Foote, J.~Leibs, R.~Wheeler,
  A.~Y. Ng \emph{et~al.}, ``Ros: an open-source robot operating system,'' in
  \emph{ICRA workshop on open source software}, vol.~3, no. 3.2.\hskip 1em plus
  0.5em minus 0.4em\relax Kobe, Japan, 2009, p.~5.

\end{thebibliography}
\bibliographystyle{IEEEtran}
}

\end{document}